\documentclass[11pt, a4paper, twocolumn]{berkeley}

\usepackage[all]{hypcap}
\usepackage{xspace}
\usepackage{subcaption}
\usepackage{wrapfig}
\usepackage{lipsum}

\makeatletter
\def\mathcolor#1#{\@mathcolor{#1}}
\def\@mathcolor#1#2#3{%
  \protect\leavevmode
  \begingroup
    \color#1{#2}#3%
  \endgroup
}
\makeatother

\usepackage{microtype}
\usepackage{graphicx}
\usepackage{booktabs} %

\usepackage{hyperref}

\usepackage{amsmath,amsfonts,bm}

\def\eqref#1{equation~\ref{#1}}

\def\1{\bm{1}}

\DeclareMathAlphabet{\mathsfit}{\encodingdefault}{\sfdefault}{m}{sl}
\SetMathAlphabet{\mathsfit}{bold}{\encodingdefault}{\sfdefault}{bx}{n}

\def\gA{{\mathcal{A}}}

\def\gD{{\mathcal{D}}}

\def\gL{{\mathcal{L}}}
\def\gM{{\mathcal{M}}}

\def\gS{{\mathcal{S}}}

\def\gZ{{\mathcal{Z}}}

\newcommand{\E}{\mathbb{E}}

\newcommand{\R}{\mathbb{R}}

\usepackage[utf8]{inputenc}
\usepackage[T1]{fontenc}    %
\usepackage{hyperref}       %
\usepackage{url}            %
\usepackage{booktabs}       %
\usepackage{amsfonts}       %
\usepackage{nicefrac}       %
\usepackage{microtype}      %
\usepackage{multicol,lipsum}
\usepackage[skip=2pt,font=footnotesize]{caption}
\usepackage{comment}
\usepackage{booktabs, tabularx} %
\usepackage{times}
\usepackage{url}
\usepackage{algorithm}
\usepackage{algorithmic}
\usepackage{graphicx}
\usepackage{mathtools}
\usepackage{amsmath}
\usepackage{amssymb}
\usepackage{amsthm} 
\usepackage{wrapfig}
\usepackage{thmtools}
\usepackage[capitalise,nameinlink,noabbrev]{cleveref}
\usepackage{enumitem}
\usepackage{float}
\usepackage[export]{adjustbox}
\usepackage{xcolor,colortbl}
\usepackage{amsfonts}
\usepackage{bm}
\usepackage{xpatch}
\usepackage{xspace}
\usepackage{listings}
\usepackage[symbol]{footmisc}
\usepackage{pifont}%

\DeclarePairedDelimiterX{\inp}[2]{\langle}{\rangle}{#1, #2}
\newtheorem{remark}{Remark}

\newlength\savewidth

\usepackage{xifthen}

\newtheorem{proposition}{Proposition}

\newcommand{\app}{\raise.17ex\hbox{$\scriptstyle\sim$}}

\usepackage{pifont}%

\usepackage{nicefrac,xfrac}
\usepackage{xcolor}
\usepackage{soul}

\definecolor{baselinecolor}{gray}{.9}

\definecolor{deemph}{gray}{0.6}

\usepackage{tabulary}
\newcolumntype{x}[1]{>{\centering\arraybackslash}p{#1pt}}

\usepackage{amsmath}
\usepackage{amssymb}
\usepackage{mathtools}
\usepackage{amsthm}
\usepackage{float}
\usepackage[font=small,labelfont=bf,tableposition=top]{caption}

\usepackage[textsize=tiny]{todonotes}

\title{Reinforcement Learning from Passive Data \\ via Latent Intentions}
\runningtitle{Reinforcement Learning from Passive Data via Latent Intentions}

\reportnumber{} %

\renewcommand{\today}{April 2023} 

\author[ ]{Dibya Ghosh}
\author[ ]{Chethan Bhateja}
\author[ ]{Sergey Levine}

\affil[ ]{UC Berkeley}

\correspondingauthor{Dibya Ghosh (\href{mailto:dibya@berkeley.edu}{dibya@berkeley.edu})}

\begin{document}

\let\oldtwocolumn\twocolumn
\renewcommand\twocolumn[1][]{%
    \oldtwocolumn[{#1}{
    \begin{center}
    \vspace{.5cm}
    \includegraphics[width=\linewidth]{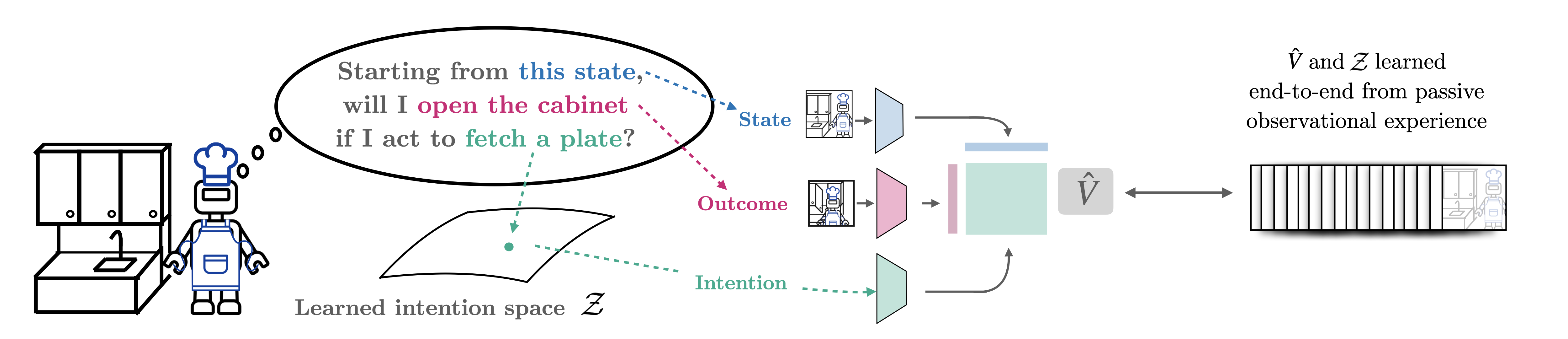}

           \captionof{figure}{We seek to extract a general knowledge of how an agent may act to influence its environment by pre-training on passive data. Our approach models the effects of acting with intention: jointly learning a latent space of agent intentions and an intention-conditioned value function that estimates the likelihood of witnessing any given outcome in the future when acting according to some latent intention. }
    \label{fig:main_fig}
    \vspace{.5cm}
        \end{center}
    }]
}

\begin{abstract}
Passive observational data, such as human videos, is abundant and rich in information, yet remains largely untapped by current RL methods. 
Perhaps surprisingly, we show that passive data, despite not having reward or action labels, can still be used to learn features that accelerate downstream RL. Our approach learns from passive data by modeling intentions: measuring how the likelihood of future outcomes change when the agent acts to achieve a particular task.
We propose a temporal difference learning objective to learn about intentions, resulting in an algorithm similar to conventional RL, but which learns entirely from passive data. When optimizing this objective, our agent simultaneously learns representations of states, of policies, and of possible outcomes in an environment, all from raw observational data. Both theoretically and empirically, this scheme learns features amenable for value prediction for downstream tasks, and our experiments demonstrate the ability to learn from many forms of passive data, including cross-embodiment video data and YouTube videos.

\end{abstract}
\maketitle

\section{Introduction}

In many reinforcement learning (RL) domains, there is a sea of data untapped by our current algorithms. For human-assistive robots, there are thousands of hours of videos of humans doing chores; for dialogue agents, expansive web corpora of humans conversing with each other; for self-driving cars, countless hours of logs of human drivers.  Unfortunately, this data cannot directly be ingested by conventional RL methods, as it often depicts agents with different embodiments, with different goals, and in different scenarios from the agent we wish to control.
Nonetheless, such datasets are diverse and abundant, and might offer a path for RL agents to learn more efficiently and generalize more broadly.

How can an RL agent learn from passive observational experience, such as videos of humans? When learning from active data, the learning signal for most RL methods comes from modeling the value function, which measures how the expected cumulative reward changes when a counterfactual action is taken. With the absence of actions and rewards in passive data, it is neither clear what quantity should be estimated (in lieu of future rewards) nor what counterfactuals should be estimated (in lieu of actions).

In this work, we propose to replace both of these missing components with a learned representation
of intentions, where intentions correspond to different outcomes a policy may seek to achieve. Rather than estimating the counterfactual effect of taking an action, we can model the effect of following a counterfactual intention from a given state; in lieu of predicting future rewards, we can estimate the likelihood of achieving any given outcome. By leveraging passive data to learn the effects of a diverse range of intentions, we can learn state representations useful for modeling the value function for any task an agent may seek to optimize downstream using active data. 

To instantiate this idea, we train an \emph{intention-conditioned value function} $V(s, s_+, z)$, which models the likelihood of achieving some future outcome $s_+$, when we start at state $s$ and act according to some latent intention $z$.
This mimics the conventional value function in RL, and we show how it may similarly be trained using temporal-difference objectives on passive data. We estimate this value using a multi-linear model that is linear in a representation of state $\phi(s)$, a representation of future outcome $\psi(s_+)$, and representation of intention. We prove that if perfectly optimized, the learned state representations can represent optimal value functions for any reward function.

The primary contribution of this paper is a pre-training scheme for RL from passive data, which models the effect of agents that act with various intentions or seek to realize various objectives.
Our approach learns from passive data
via RL-like temporal difference backups utilizing two main ideas, prediction of possible outcomes (future states) as the learning objective in lieu of rewards, and the use of learned latent representations of intentions in lieu of actions. We show both formally and empirically that the representations learned through pre-training on passive data  for representing downstream RL value functions. Our experiments demonstrate the ability to learn useful features for downstream RL from a number of passive data sources, including learning from videos of different embodiments in the XMagical benchmark, and learning from raw YouTube videos for the offline Atari benchmark.

\section{Problem Formulation}

We study pre-training from passive observational experience to improve the performance of downstream RL. We formalize this problem in the notation of Markov control processes, environments parameterized as $\gM \coloneqq (\gS, \gA, P, \rho)$ with state space $\gS$, action space $\gA$, transition kernel $P(s'|s, a)$, and initial state distribution $\rho(s_0)$.

We consider a two-stage process: a pre-training stage in which the agent trains only on passive data, followed by a downstream RL stage in which the agent must solve some specified task. The goal during the pre-training phase will be to acquire state representations generally useful for future tasks that may be assigned downstream to the agent. 

During this pre-training phase, the learning algorithm gets access to sequences of observations $\left((s_{0}^i, s_{1}^i, \dots, )\right)_{i=1}^n$ collected by some behavioral policy $\pi_{\beta}(a|s)$ acting in an Markov control process $\gM$. In comparison to the typical ``active'' experience, this passive data cannot be directly used for an RL task, since it lacks both action annotations and reward signal. Rather, we use this passive data to learn a state representation $\phi(s) \in \R^d$ for a downstream RL agent.

Downstream, the algorithm is tasked with learning a policy $\pi(a|s)$ that acts to maximize expected discounted return: $\E_{\pi}[\sum_{t} \gamma^t r(s_t)]$ for some task reward $r(s)$. For simplicity in exposition, we will assume that the downstream agent acts in the same MDP $\gM$, but in our experiments, we will show that our method also works when the action space and transition dynamics deviate from the environment used to collect the passive data.
While a state representation $\phi(s)$ learned on passive data may be used in many ways for downstream RL (exploration bonuses, curriculum learning), we focus on using the learned representation as a feature basis, for example for estimating the agent's value function.

\section{Reinforcement Learning using Passive Data}

We would like to pre-train on passive experience in a way that results in useful features for solving downstream RL tasks. Perhaps surprisingly, we can learn such features by running a form of ``reinforcement learning'' on the passive data, despite this data having neither reward nor action labels. We will see that rewards and actions can be replaced with more generic abstractions that are more amenable to the passive data setting. %

When learning from active experiential data, conventional value-based RL methods learn by modeling the state-action value function \mbox{$Q(s,a) = \E[\sum_{t=0}^\infty \gamma^t r(s_t) | s_0=s, a_0=a]$},
the future discounted return of a policy %
after following an action $a$ for one time-step from the state $s$. The $Q$-function embeds knowledge about upcoming rewards and how they vary depending on the counterfactual action chosen, relevant quantities when acting to maximize accumulated reward. 

In the absence of rewards and actions, we can generalize the state-action value function through the use of \textit{intentions} and \textit{outcomes}. An intention $z$ corresponds to a policy acting to optimize a specific objective, for example maximizing some pseudo-reward \citep{Sutton2011HordeAS} or reaching a specific goal state \citep{Schaul2015UniversalVF}. Just like actions, we can use intentions as counterfactual queries, by asking how the environment would change if we acted to optimize the objective corresponding to the intention $z$.

Formally, we define $\gZ$ to be a space of intentions, where each intention $z$ corresponds to an reward function $r_z: s \mapsto r$ that defines the objective for an agent pursuing intention $z$. An agent acts according to intention $z$ when the agent's induced state transition distribution corresponds to acting optimally to maximize expected cumulative pseudo-reward $r_z(s)$, which we write as $P_z(s'|s) \coloneqq P^{\pi_{r_z}^*}(s'|s)$).
The \textbf{intention-conditioned value function (ICVF)} is defined as 
\begin{equation}
\label{eq:icvf}
V(s, s_+, z) =  \E_{s_{t+1} \sim P_z(\cdot|s_t)}\left[\sum\nolimits_{t} \gamma^t 1(s_t=s_+) | s_0 = s\right],
\end{equation}
the un-normalized likelihood of witnessing an outcome $s_+ \in \gS$ in the future when the agent acts according to intention $z$ starting from a state $s$. %

The ICVF answers queries of the form ``How likely am I to see \underline{~~~~~~~~} if I act to do \underline{~~~~~~~~~} from this state?''. This generalizes both rewards and actions in a conventional value function; rather than predicting expected future reward, we instead predict the chance of seeing a future outcome. Instead of estimating the counterfactual ``what happens in the environment when I take an action $a$'', we instead model the counterfactual ``what happens in the environment when I follow an intention $z$''. {\let\thefootnote\relax\footnote{{See Appendix \ref{appendix:icvf_example} for didactic visualizations of the ICVF in a gridworld domain.}}}

The ICVF forms a successor representation \citep{Dayan1993ImprovingGF}, which means that although it only models the likelihood of seeing future states $s_+$, the function can be used to compute expected return for any reward signal.

\begin{remark}
    \label{eq:icvf_representation}
    For any reward signal $r: \gS \to \R$, we can express $V_r^z(s) =\E_{P_z}[\sum\nolimits_{t} \gamma^t r(s_t) | s_0=s]$, the future expected return when acting with intent $z$ using the ICVF:
    \begin{equation}
        V_r^z(s) = \sum_{s_+ \in \gS} r(s_+)V(s, s_+, z)
    \end{equation}    
\end{remark}

This property means that intention-conditioned value functions describe not only how to achieve specific goal outcomes, but rather more generally performing \textit{any task} in an environment. This generality, combined with the fact that ICVFs are not tied to any action abstraction, makes the ICVF an appealing target for pre-training from passive observational experience. Temporal abstractness is particularly important when learning from passive data; even if the downstream agent's dynamics do not match the passive data locally, the 
 general relationships between states, future outcomes, and possible intentions encoded by an ICVF may still transfer downstream.

\section{Learning Representations from ICVFs}

The intention-conditioned value function encodes general knowledge about how agents may behave in an environment, but as a black-box function of three inputs, it is not obvious how this knowledge should be used downstream. We consider attempting to extract the knowledge in an ICVF into a state representation $\phi(s)$ to provide for a downstream RL task. We seek two desiderata: 1) state representations should be easy to extract from our learned model, and 2) these state representations should be useful for representing value functions of task rewards we might face downstream.

We motivate our approach by investigating the structure underlying an intention-conditioned value function. Since an ICVF describes a successor representation, we can write the value for a state $s$, intention $z$ and future outcome $s_+$ in vector notation as  
$$V(s, s_+, z) = e_{s}^\top (I - \gamma P_{z})^{-1} e_{s_+},$$
where $(I-\gamma P_{z})^{-1} = M_{z}$ is the successor representation \citep{Dayan1993ImprovingGF} for the agent acting according to the intention $z$, and $e_{s}, e_{s+}$ refer to standard basis vectors in $\R^{|\gS|}$. This means that the ICVF is linear separately in three inputs: a state representation $\phi(s) = e_{s}$, an outcome representation $\psi(s_+) = e_{s_+}$, and a representation of the counterfactual intention $T(z) = M_{z}$. We propose to learn an approximation that also respects this multilinear structure, letting
$$\hat{V}_\theta(s, s_+, z) = \phi_\theta(s)^\top T_\theta(z) \psi_\theta(s_+),$$
where we replace the three canonical representations with learned latent representations: \mbox{$e_s$ with $\phi_\theta(s) \in \R^d$}, \mbox{$e_{s_+}$ with $\psi_\theta(s_+) \in \R^d$} and \mbox{$M_z$ with $T_\theta(z) \in \R^{d \times d}$}. 

Perhaps surprisingly, the state representation $\phi(s)$ will satisfy both criterion we laid out. The first is met by construction, since we are now explicitly learning a state representation $\phi(s)$ that can be passed to a downstream RL algorithm. For the second, we will show that if our multilinear model can learn a good approximation to the ICVF, then our learned state representation $\phi(s)$ will be able to linearly approximate value functions for any downstream task.

\begin{proposition}[Downstream value approximation]
    Suppose $\phi, \psi, T$ form an approximation to the true ICVF with error $\epsilon$, that is $\forall z \in \gZ$,
    $$\sum_{s, s_+ \in \gS}(V(s, s_+, z) - \phi(s)^\top T(z) \psi(s_+))^2 \leq \epsilon.$$
    For all rewards $r(s)$ and intentions $z \in \gZ$, $\exists \theta_r^z \in \R^d$ s.t.
    \begin{equation}\sum_{s \in \gS} (V_r^z(s) - \phi(s)^\top \theta_r^z)^2 \leq \epsilon \sum_{s_+ \in \gS} r(s_+)^2.
    \end{equation}
\end{proposition}

This statement connects accuracy in modelling the ICVF for an intention set $\gZ$ to the accuracy to which we can modelling downstream value functions when acting to pursue these intentions. Intuitively, if we can use passive data to learn a multilinear ICVF for a large set of intentions, we effectively learn a state representation equipped to properly estimate downstream value functions we may be tasked to model.

Why do the learned representations have this nice property?  Consider the simpler setting where we are able to learn $\phi$, $\psi$, and $T$ with no approximation error, that is \mbox{$V(s, s_+, z) = \phi(s)^\top T(z) \psi(s_+)$}. Since downstream value functions $V_r^z$ can be expressed using an ICVF (Remark \ref{eq:icvf_representation}), we can decompose the downstream value function as,
\begin{align}
V_r^z(s) &= \sum\nolimits_{s_+ \in \gS}r(s_+)V(s, s_+, z)\\
&= \phi(s)^\top T(z) (\sum\nolimits_{s_+ \in \gS} r(s_+)\psi(s_+)).
\end{align}
Overloading notation $\psi(r) = \sum\nolimits_{s_+ \in \gS} r(s_+)\psi(s_+)$,
\begin{align}
V_r^z(s) &= \phi(s)^\top T(z) \psi(r).
\end{align}

This proves the desired result, that downstream value functions can be expressed as linear functions of $\phi(s)$ (using parameter \mbox{$\theta_r^z = T(z) \psi(r)$}). 

Looking closer at the derivation reveals a few interesting details about the multi-linear ICVF.  While $\psi$ is explicitly defined as a representation of future outcome states (i.e. of $\psi(s_+)$), through the overloading, it \textit{implicitly} also provides a latent representation for reward functions $\psi(r)$. That is, when learning an multi-linear ICVF, we are concurrently learning a \textit{space of latent intentions}, where reward functions $r_z: \gS \to \R$ get mapped to latent intentions $z = \hat{\psi}(r_z) \in \R^d$. The formal properties of such a learned latent intention space are studied in greater depth by \citet{Touati2021LearningOR}. A second consequence is that our ICVF approximation defines an estimate for $V_z^*(s)$, the optimal value function for an intent $z$: \mbox{$V_z^*(s) \approx \hat{\phi}(s)^\top \hat{T}(z)\hat{\psi}(r_z)$}, which we write henceforth as $\hat{V}(s, z, z)$. As we will see in the next section, these estimates will play a role in a temporal difference learning loop to learn an ICVF from passive data.

\begin{algorithm*}
\caption{Learning Intent-Conditioned Value Functions from Passive Data}\label{alg:cap}
\begin{algorithmic}
\STATE Receive passive dataset of observation sequences $\gD = \left((s_{0}^i, s_{1}^i, \dots,) \right)_{i=1}^n$
\STATE Choose intention set $\gZ$ to model, e.g. the set of goal-reaching tasks
\STATE Initialize networks $\phi: \gS \to \R^d$, $\psi: \gS \to \R^d$,  $T: \gZ \to \R^{d \times d}$
\STATE Define ICVF $V_\theta(s, s_+, z) = \phi(s)^\top T(z) \psi(s_+)$ and derived value model $V_\theta(s, z, z) = \phi(s)^\top T(z)\psi(r_z)$
\REPEAT
\STATE Sample transition $(s, s') \sim \gD$, potential future outcome $s_+ \sim \gD$, intent $z \in \gZ$
\STATE Determine whether transition $s \leadsto s'$ corresponds to acting with intent $z$ by measuring advantage
$$A = r_z(s) + \gamma V_\theta(s', z, z) - V_\theta(s, z, z)$$
\STATE Regress $V_\theta(s, s_+, z)$ to $1(s=s_+) + \gamma V_{target}(s', s_+, z)$ when advantage of $s \leadsto s'$ is high under intent $z$
$$\gL(V_\theta) = \E_{(s, s'), z, s_+}[|\alpha - 1(A < 0)|\left(V_\theta(s, s_+, z) - 1(s=s_+) - \gamma V_{target}(s', s_+, z)\right)^2]$$
\UNTIL{convergence}
\STATE Return $\phi(s)$ as a state representation for use in downstream RL
\end{algorithmic}
\end{algorithm*}

\section{Learning ICVFs from Passive Data}

Having established learning (multi-linear) intention-conditioned value functions as a useful pre-training objective for downstream RL, we now develop a practical method for learning these value functions from passive data. Our method learns a multi-linear ICVF via temporal difference learning; the resulting process looks very similar to a standard RL algorithm, and indeed our approach to pre-training on passive data might be best described as an RL approach to representation learning.

We begin our derivation assuming that we have access to a dataset of action-labelled transition data $\gD = \{(s, a, s')\}$. Later, we will drop the dependence on actions, making it compatible with passive data. The general recipe for training value functions is to perform a temporal difference update towards a fixed point, which for ICVFs is defined as :
\begin{remark}
The true ICVF satisfies the following Bellman equation for all $s, s_+ \in \gS, z \in \gZ$:     
$$V(s, s_+, z) = \E_{a \sim \pi_z^*}[1(s=s_+) + \gamma \E_{s'}[V(s', s_+, z)]],$$

where $\small \pi_z^* = \arg\max_{a} r_z(s) + \gamma \E_{s'}[V(s', z, z)]$. 
\end{remark}

This form of fixed-point motivates the training of ICVFs using value-learning methods stemming from value iteration.
We choose to adapt IQL \citep{kostrikov2021offline}, a simple and empirically stable value learning objective which softens the maximum operator into that of an expectile. For an expectile $\alpha \in [\frac{1}{2}, 1]$, this yields the following objective for any intent $z \in \gZ$ and future outcome $s_+ \in \gZ$:
\begin{multline}
\label{eq:loss}
    \small\min \E_{(s, a, s') \sim \gD}[\underbrace{|\alpha - 1(A < 0)|}_{\text{weighting}}(\underbrace{V_\theta(s, s_+, z)}_{\text{current estimate}} -  \\~~~~\small \underbrace{1(s=s_+) - \gamma V_{target}(s', s_+,  z)}_{\text{target estimate}})^2]
\end{multline}
where $A = r_z(s) + \gamma \E_{s'\sim P(s, a)}[V(s', z, z)] - V(s, z, z)$ is the current estimated advantage of $s \leadsto s'$ when acting according to intention $z$. This expression re-weighs the TD-learning backup to be focus more on state transitions that are favorable under $z$ (when $A > 0$), and to ignore state transitions that do not correspond to acting optimally to $z$. This update inherits the same convergence guarantees as IQL, meaning the learned value converges to the true value  when the expectile is taken to the limit $\alpha \to 1$.

Notice that this objective depends mainly on $s$ and $s'$, with the only dependency on $a$ in the expected next state value in the definition of $A$. To make the objective action-free, we propose to estimate this expectation with the single sample estimator:
\mbox{$\hat{A} = r_z(s) + \gamma V(s', z, z) - V(s, z, z)$}. Roughly, this approximation estimates whether a transition $(s, s')$ is likely to have been generated by following an intention $z$, by measuring whether the predicted value for $z$ is higher at the new state $s'$ than it was at the original. The single-sample approximation is unbiased under deterministic dynamics and exogenous noise but known to be susceptible to an optimism bias in stochastic environments \citep{Paster2022YouCC}. In our experiments, we found this bias to not be a major nuisance, but we note that other (more complicated) techniques can correct the biased approximation even under stochastic dynamics \citep{Yang2022DichotomyOC, pmlr-v162-villaflor22a}.

\textbf{Summary:} As a representation learning objective, we learn a multi-linear intent-conditioned value function $V_\theta(s, s_+, z)$ from a passive dataset of observation transitions $\gD = \{(s, s')\}$. To instantiate the algorithm, we must choose a set of intentions to model. In our experiments, we choose the generic set of goal-reaching tasks in the dataset, so each intention corresponds to a reward $r_z(s) = 1(s=s_z)$ for some desired state $s_z \in \gD$. Since the outcome representation $\psi(\cdot)$ we learn also implicitly defines a latent representation of intentions, we can associate the task of reaching $s_z$ with the latent intention $z = \psi(r_z) \coloneqq \psi(s_z)$. 

We train our multi-linear ICVF end-to-end using a temporal-difference learning objective based on the IQL value update (see Algorithm \ref{alg:cap}). Each step, the method samples a transition $(s, s')$, an arbitrary future outcome state $s_+$, and an arbitrary desired outcome state $s_z$ (intention $z = \psi(s_z)$), and optimizes Equation \ref{eq:loss}. This objective uses the advantage estimator to determine whether $s \leadsto s'$ corresponds to acting optimally under the intention $z$. If deemed so, the estimate at the next state $V(s', s_+, z)$ is used to bootstrap the estimate for the current state $V(s, s_+, z)$. After training has converged, the learned state representation $\phi$ is extracted and passed onto a downstream RL algorithm to use as a feature basis for downstream value learning.

\section{Related Work}

\textbf{Learning from passive data:} A number of prior works have approached pre-training on prior data as a problem of representation learning or prediction. The former leads to reconstruction objectives 
with autoencoders \citep{Nair2018VisualRL, Seo2022MaskedWM, Xiao2022MaskedVP} or augmentation-driven contrastive objectives \citep{Srinivas2020CURLCU}. Such objectives are general and widely used but do not capture any information about environment dynamics. In the latter category, prior works have proposed objectives that pair current states with future ones via contrastive learning \citep{Sermanet2017TimeContrastiveNS, Nair2022R3MAU}, predict the next state from the current one \citep{Seo2022ReinforcementLW}, or learn distances between states and goals \citep{Ma2022VIPTU}.
Among these approaches, \citet{Ma2022VIPTU} also learns a value function, but differs in that they do not model arbitrary counterfactual outcomes and use BRM over the more conventional temporal-difference learning.
These methods fall under the umbrella of behavioral prediction, modeling the future of the environment when the agent acts according to the policy used to collect the passive data. In contrast, our method models environment futures for many different policies with different intentions, not just the data-collection policy. Modeling a range of intentions allows us to capture more control-relevant features, and we verify in our experiments that ICVF representations lead to improved downstream RL over these approaches.

In lieu of pre-training, another approach is to train on passive data alongside incoming active experience during the downstream task. In third-person imitation learning, agents learn from videos of expert demonstrations, for example by adversarially training to distinguish between states in passive and active data \citep{Stadie2017ThirdPersonIL, Torabi2018GenerativeAI} or direct reward learning \citep{Edwards2019PerceptualVF}. Another common approach is to train an inverse model on active agent data, and use it to generate fake action and reward annotations for the passive data for downstream RL \citep{Torabi2018BehavioralCF, Schmeckpeper2020ReinforcementLW, Baker2022VideoP, Chang2022LearningVF}. This limits the agent to learning only from passive data similar to that of the downstream agent, as otherwise, the inverse model cannot generate appropriate labels. Other approaches train value functions by synthesizing In general, joint-training is oft limited by the amount of active experiental data, and the computational requirements downstream scale poorly with passive data size compared to pre-training approaches.

\textbf{Successor features and GCRL:} Our intent-conditioned value function is conceptually related to successor representations and universal value functions. The successor representation encodes environment futures under a fixed policy, from the original tabular setting \citep{Dayan1993ImprovingGF} to more general function approximation schemes \cite{Barreto2016SuccessorFF, Kulkarni2016DeepSR, Machado2017EigenoptionDT}. Universal value functions represent value functions for many policies, such as a pre-defined set \citep{Sutton2011HordeAS} or the set of optimal goal-reaching policies \citep{Schaul2015UniversalVF}. \citet{Borsa2018UniversalSF} and \citet{Touati2021LearningOR} combine these two ideas to learn the value of many rewards for many policies. Their approaches are analogues to the ICVF in the active experiential data setting. Our multilinear representation is closest to the bilinear one proposed by \citet{Touati2021LearningOR}, who use active data to learn a joint representation of states, actions, and intents, $F(s, a, z)$ and a representation of outcomes $B(s_+)$.

\section{Experiments}

Our evaluation studies ICVFs as a pre-training mechanism on passive data, focusing on the following questions:

\begin{itemize}[noitemsep]
\item Can we learn ICVFs from passive data?
\item How well can the extracted representations represent value functions for downstream RL tasks?
\item How do our representations compare to other approaches for pre-training on passive data?
\end{itemize}

To encompass the many forms of passive data that we may wish to pre-train on, we evaluate on passive data from the D4RL benchmark \citep{Fu2020D4RLDF}, videos of agents with different embodiments in the XMagical benchmark \citep{toyer2020magical,zakka2021xirl}, and scraped Youtube videos of Atari 2600 games \citep{bellemare13arcade}. Accompanying code can be found at \href{https://github.com/dibyaghosh/icvf_release}{https://github.com/dibyaghosh/icvf\_release}

\begin{figure*}
    \centering
    \includegraphics[width=0.28\linewidth]{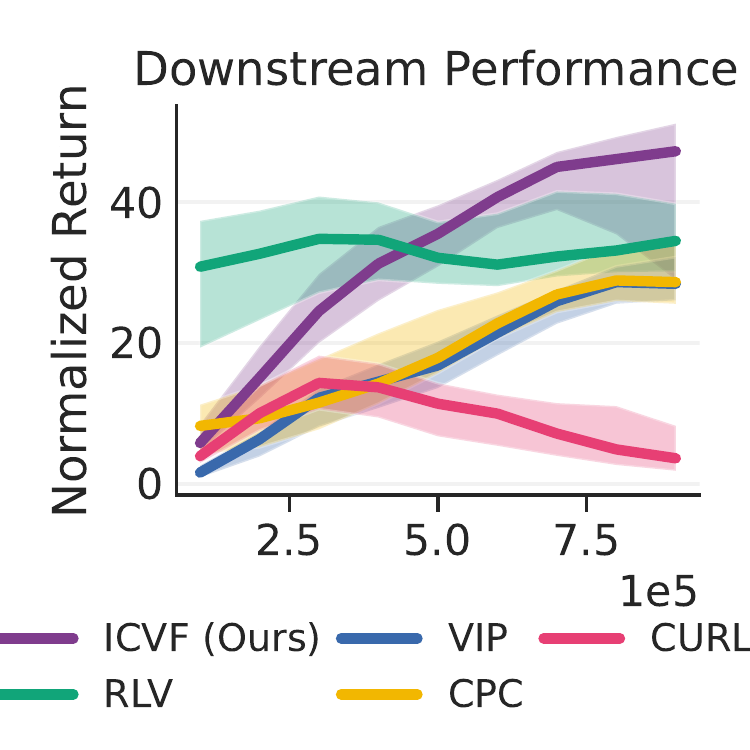}
    \includegraphics[width=0.28\linewidth]{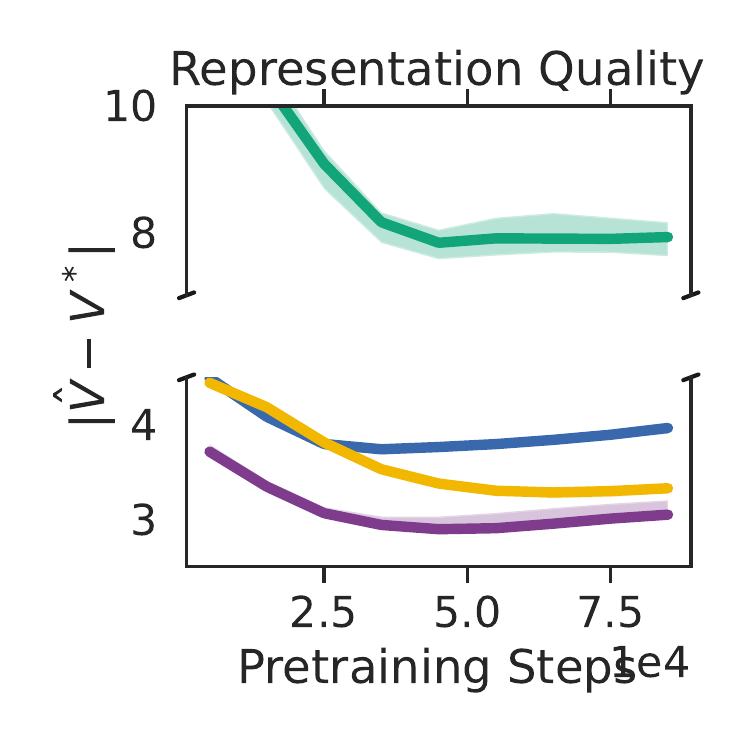}
    \includegraphics[width=0.28\linewidth]{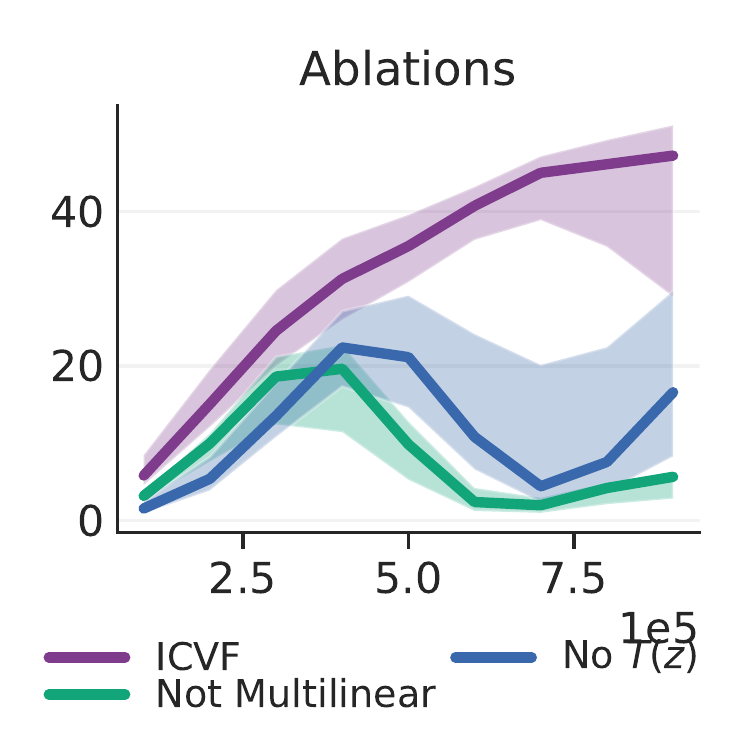}
    \caption{\textbf{D4RL Antmaze tasks:} We compare ICVF representations pre-trained on passive data from \texttt{antmaze-large-diverse-v2} to representations from other methods for downstream RL performance (left) and for fidelity of value approximations (middle). Ablating our approach, we find both components of our ICVF method to be important for performance: learning future outcomes from many intents, and the use of the multi-linear decomposition (middle). (Additional Antmaze tasks in Table \ref{tab:antmaze} and \ref{appendix:d4rl_extra}.}
    \label{fig:main_antmaze}
\end{figure*}

\subsection{Experimental setup and comparisons}

We focus on the offline RL setting, where the agent first pre-trains on a large dataset of action-less reward-less sequences of observations, before performing offline RL on a much smaller dataset of standard active experience, which includes actions and the task reward. During the finetuning phase, we run a standard value-based offline RL algorithm, initializing the parameters of the value function $V(s)$ using the learned representation and adding the representation loss as an regularizing penalty to prevent the representation from deviating far from the pre-trained one. See Appendix \ref{appendix:exp_setup} for full details about the experimental setup, including architecture and hyperparameter choices.

We compare the representations learned through the ICVF to standard representation learning methods that are compatible with learning from actionless data. We compare to CURL \citep{Srinivas2020CURLCU}, which matches state representations between two augmentations of the same versions of the same state; Contrastive RL (CPC) \citep{Oord2018RepresentationLW, Eysenbach2022ContrastiveLA}, which matches state representations with those of future states in the same trajectory; APV \citep{Seo2022ReinforcementLW}, which estimates next states using a latent stochastic model; and VIP \citep{Ma2022VIPTU}, which learns representations by estimating distances between states using a Fenchel dual value function, and has previously been proposed as a way to learn from passive data. We also compare to RLV \citep{Schmeckpeper2020ReinforcementLW}, which trains an inverse model on the downstream actionful data, and labels the passive data with fake action labels for the downstream RL algorithm to ingest. Note that the assumptions for RLV are significantly different from our method (and the other comparisons), since RLV cannot train on the passive data without already having access to actionful data. %

\subsection{Learning from the same embodiment}
\begin{table}
	\centering
	\caption{Performance of IQL trained on 250k actionful transitions on D4RL AntMaze tasks after representation pre-training on dataset of 1M observation-only transitions with each of the representation learning methods in our comparison, as well as an oracle baseline that receives action and reward labels for the entire passive dataset. Results with 5 seeds.}
	\small
	\vspace{0.2em}
	\label{tab:antmaze}
	\begin{adjustbox}{max width=\linewidth}
	\begin{tabular}{l|cc|cc}
			\toprule
			 &
			 \multicolumn{2}{|c|}{\texttt{antmaze-medium-*}}
             & 
                \multicolumn{2}{|c}{\texttt{antmaze-large-*}}\\
                & Diverse & Play & Diverse & Play \\ 
			\midrule
            No Pretraining &  0.45 {\tiny $\pm 0.05$} & 0.45 {\tiny $\pm 0.10$} & 0.18 {\tiny $\pm 0.11$} & 0.26 {\tiny $\pm 0.07$} \\
            CURL & 0.44 {\tiny $\pm 0.30$} & 0.22 {\tiny $\pm 0.28$} & 0.16 {\tiny $\pm 0.08$} & 0.03 {\tiny $\pm 0.03$} \\
            Contrastive RL (CPC) & 0.57 {\tiny $\pm 0.04$} & 0.42 {\tiny $\pm 0.05$} & 0.06 {\tiny $\pm 0.01$} & \textbf{0.29 {\tiny $\pm 0.02$}} \\
            RLV & 0.46 {\tiny $\pm 0.25$} & \textbf{0.75 {\tiny $\pm 0.05$}} & \textbf{0.33 {\tiny $\pm 0.03$}} & \textbf{0.35 {\tiny $\pm 0.07$}} \\
            \textbf{ICVF (ours)} & \textbf{0.66 {\tiny $\pm 0.06$}} & \textbf{0.75 {\tiny $\pm 0.02$}} & \textbf{0.32 {\tiny $\pm 0.11$}} & \textbf{0.35 {\tiny $\pm 0.02$}} \\
            \midrule
            Oracle & 0.75 {\tiny $\pm 0.01$} & 0.75 {\tiny $\pm 0.06$} & 0.42 {\tiny $\pm 0.07$} & 0.41 {\tiny $\pm 0.03$} \\
            
\bottomrule
		\end{tabular}
	\end{adjustbox}
\end{table}

We first study the Antmaze tasks from the D4RL benchmark \citep{Fu2020D4RLDF}, a domain that is easy to visualize, and where optimal policies and value functions can be computed for reference. During pre-training, the agent receives a passive dataset of state-observation sequences of the agent moving to different locations in a maze  ($1 \times 10^6$ frames), which we construct by stripping all annotations from the publically available dataset. Downstream, the agent receives a smaller dataset ($2.5 \times 10^5$ transitions) with reward and action information, and must learn a policy to reach the top-right corner of the maze. %

To understand how well the ICVF state representation can represent the optimal value function for the downstream task, we train a linear probe on the representation to regress to $V^*(s)$ given the representation as input (errors visualized in Figure \ref{fig:main_antmaze}). The ICVF representations reduces value estimation error more than any of the alternative representations in our comparison, although CPC also performs competitively. This also translates to better downstream performance of the learned RL policy: in all four evaluated settings, our approach leads to the highest final performance (Table \ref{tab:antmaze}), in some cases with performance similar to an oracle baseline that does has access to the true action and reward labels for the entire passive dataset. RLV is also competitive for this domain. This is to be expected since the active data comes from exactly the same distribution as the passive data, the ideal case for such an inverse model.

To understand what components of the ICVF lead to improved downstream performance, we ablate the two core components: the modelling of many different intents, and the representation of the ICVF using a multi-linear approximation. When we model only a single intent (the behavioral policy generating the dataset), performance decreases on the \texttt{antmaze-large-*} tasks, but not on the \texttt{antmaze-medium-*} tasks; we hypothesize this is because it is harder to capture long-horizon reasoning in more complex environments when not modelling intent-driven behavior. In our second ablation, we compare to representing the ICVF as a black box function of some state representation $V(s, g, z) = f(\phi(s), g, z)$; this model is able to approximate the ICVF better, but the consequent representations are not amenable for downstream control. In Appendix \ref{appendix:d4rl_extra}, we perform a further probe of the qualities of the ICVF representation, in particular, measuring how well it can represent various optimal goal-reaching value functions, and visualizing value functions for the various intentions.

\begin{figure}
    \centering
    \includegraphics[width=\linewidth]
    {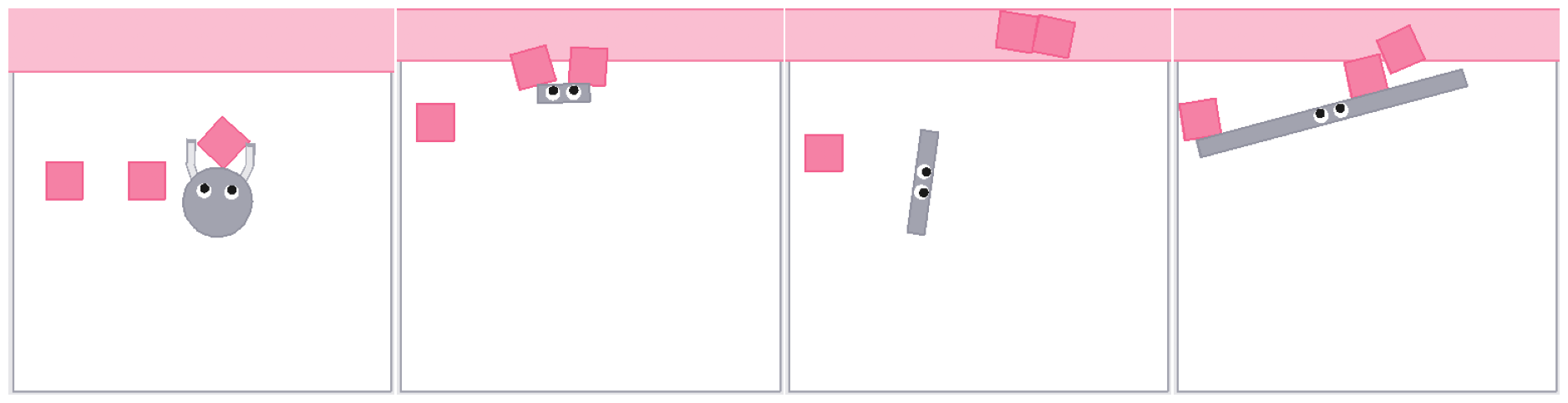}
    \includegraphics[width=\linewidth]{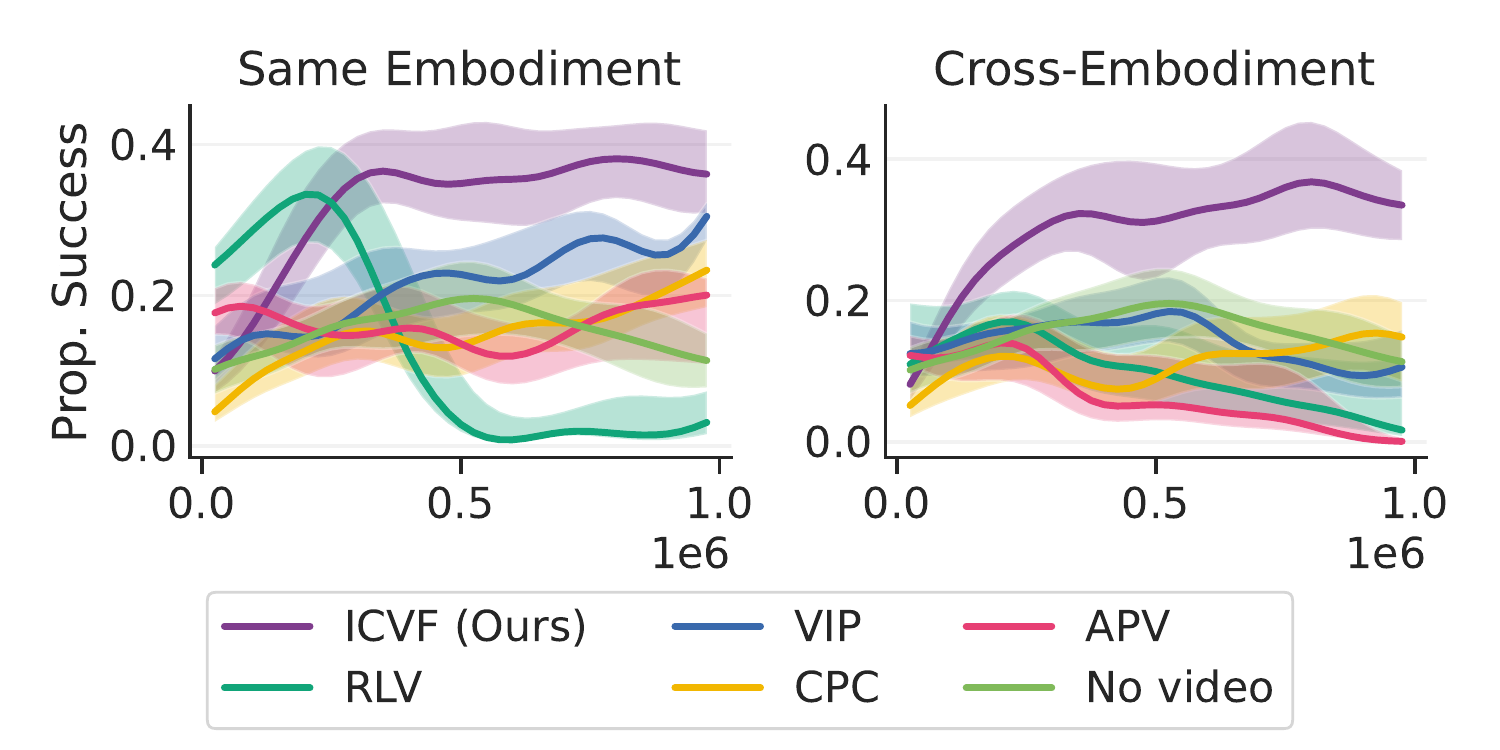}
    \caption{\textbf{XMagical}: (top) A visualization of the different agent embodiments. (bottom) Downstream performance on the Gripper embodiment after pretraining from passive same-embodiment or cross-embodiment data. ICVF representations lead to the highest downstream performance amongst all comparisons.  }
    \label{fig:xmagical}
\end{figure}

\subsection{Learning from different embodiments}

In many settings, passive experience comes from an agent with a different embodiment (for example, a robot learning from human videos). To understand how ICVFs handle such \emph{embodiment gaps}, we evaluate our method on tasks from XMagical, a benchmark for third-person imitation \citep{zakka2021xirl, toyer2020magical}. This domain has four types of agents, with differing shapes, appearances, and action spaces (see Figure \ref{fig:xmagical}). In all cases, agents receive a top-down image as observation, and are tasked with maneuvering the three blocks to the goal zone indicated in pink, receiving partial reward for each item cleared. Using the video datasets released by \citet{zakka2021xirl}, we evaluate two settings: the same-embodiment setting and the cross-embodiment setting, where the passive data consists of videos of only other embodiments.

Even when learning from images, we find that our algorithm is able to learn ICVFs from the video dataset that lead to good downstream performance. We find that methods like VIP and CPC are able to learn useful representations from videos of the same embodiment, but only ICVFs are able to learn representations that lead to good performance from cross-embodied videos. Investigating the representations learned by contrastive learning or next-state prediction, we see that they occasionally fail to focus on the position of the agent, leading to poor control performance.
\begin{figure}
    \centering
    \includegraphics[width=\linewidth]{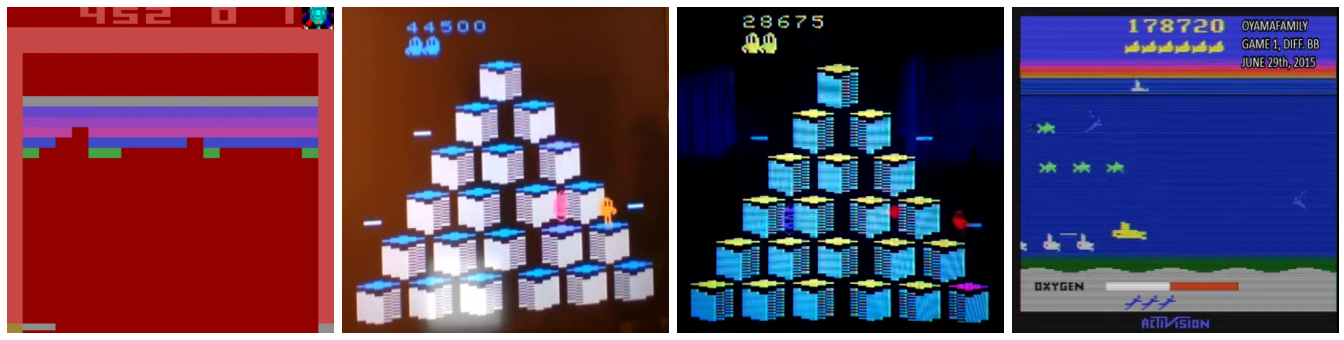}
    \caption{YouTube videos of Atari games include corruptions such as color and angle shifts, lighting differences, and text overlays.}
    \label{fig:corruptions}
\end{figure}

We note that RLV, which re-labels passive data with an inverse model, performs decently from video of the same embodiment, but fails completely when training on video of different embodiments, indicating the deficiencies of such approaches when learning from diverse sources of passive data.
\begin{figure*}
    \centering
    \includegraphics[width=\linewidth]{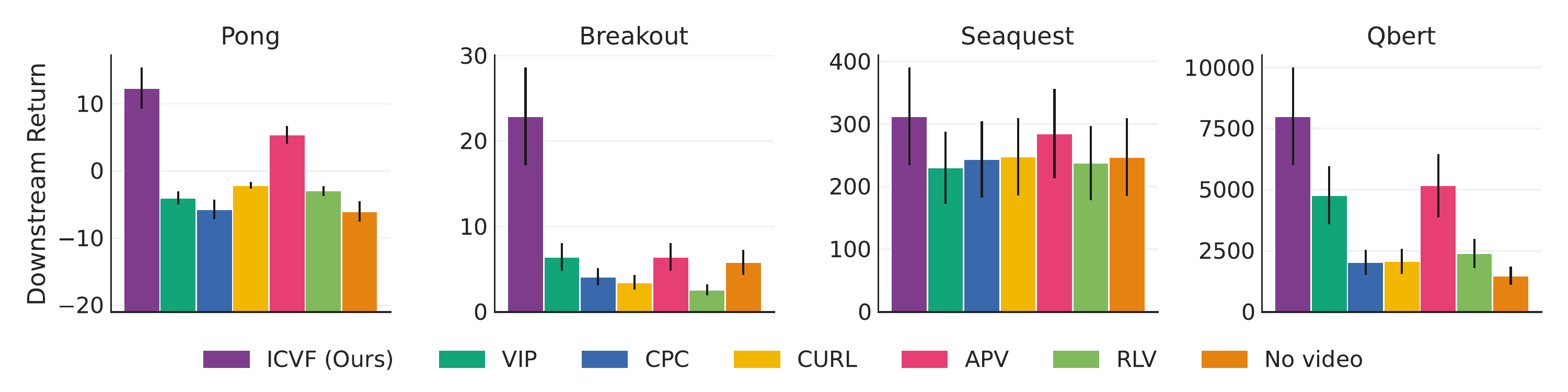}
    \caption{\textbf{Atari 2600 with YouTube Videos}: Final performance of QRDQN-CQL agent initialized with representations learned from our YouTube video dataset. In three of four games, ICVF representations lead to improved performance by large margins.}    \vspace{-1em}

    \label{fig:atari_rewards}
\end{figure*}

\subsection{Atari 2600: Learning from YouTube videos}

Finally, to test how well our approach works on more uncurated banks of passive experience, we consider pre-training for Atari 2600 games from Youtube videos. For the four Atari games evaluated by \citet{kumar2020conservative}, we scrape a dataset of videos from Youtube, with minor pre-processing (cropping, resizing, and grayscaling) to match the format of the simulated environment \citep{bellemare13arcade}. Despite this preprocessing, the Youtube videos still differ visually from the downstream simulation, as it includes data from different game modes, corruptions like color shifts, lighting differences, and text overlays , and even one video recorded with a cell-phone pointed at the screen (see Figure \ref{fig:corruptions}).
In total, we collected about 90 minutes of video for each game (exact numbers vary between games). See Appendix \ref{appendix:atari_details} for a full description of the processing pipeline and a list of Youtube videos used. Downstream, we train the CQL+QRDQN agent from \citet{kumar2020conservative} on an actionful dataset 30 minutes in length ($1 \times 10^5$ frames),
chosen by subsampling uniformly from the DQN replay dataset \citep{Agarwal2020AnOP}. We found that augmenting the Youtube dataset via random cropping and color jitters improved performance across the board, so we include it for all comparisons (details in \ref{appendix:video_augmentation}). 

\begin{figure}
    \centering
    \includegraphics[width=\linewidth]{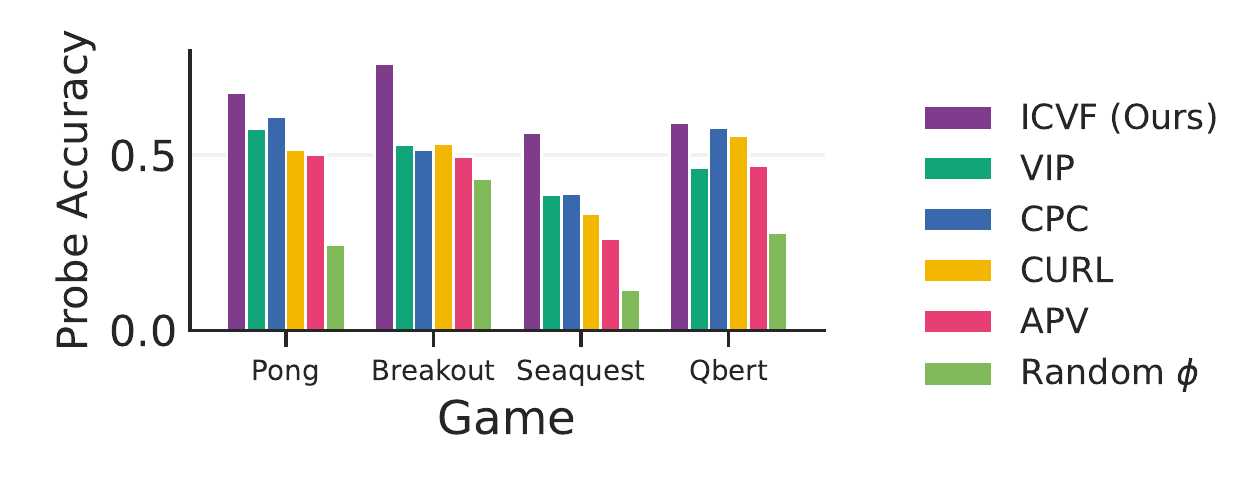}
    \caption{ICVF representations on YouTube video are better able to predict actions of an QRDQN agent trained with $100\times$ more data, as measured by a linear probe. Value probe in Appendix \ref{appendix:atari_extra}}
    \label{fig:atari_probe}
\end{figure}

We find that initializing value functions using representations learned using ICVF leads to significantly improved performance over other methods (Figure \ref{fig:atari_rewards}) on three of the four games, with ICVFs being significantly above the second highest-performing method, next-state prediction (APV). To understand whether this performance improvement manifests in other ways, we perform linear probes on the learned representations towards ``oracle'' quantities generated from an oracle agent trained with $100\times$ more data, and find that the ICVF is able to more closely model these quantities. We also visually analyze the learned ICVF (Figure \ref{fig:example_icvf} in Appendix \ref{appendix:atari_extra}); although the ICVF is not very precise locally, it correctly captures long-horizon dependencies between states, suggesting that the learned representations do capture long-horizon features.

\section{Conclusion}

In this paper, we propose to pre-train on passive datasets to model the effect of an agent acting according to different intentions. Our method jointly learns a latent space of agent intentions and an intention-conditioned value function, which estimates the likelihood of witnessing any given outcome in the future, conditional on the agent acting to realize some latent intention. We showed that this value function encodes general counterfactual knowledge of how the agent may influence an environment, and that when we learn it using a multilinear approximation, the consequent state representations are well-suited to predict downstream value functions. We design an algorithm for learning these intention-conditioned value functions from passive data, and showed that the representations learned using such objectives are well-suited for downstream control. 

Our paper constitutes one approach in understanding how RL may serve as a pretraining mechanism from passive data. From an empirical perspective, scaling the method beyond small-scale video datasets to larger corpora of human video \citep{DBLP:journals/corr/abs-2110-07058} and dialogue may lead to more useful actionable representations for downstream control. From a theoretical perspective, there is much left to understand about intention-conditioned value functions; developing more principled approaches to train them beyond our single sample heuristic, and discovering better ways of leveraging the structure of such models, may provide us with powerful ways of modeling and representing the world from passive data.

\clearpage
\section*{Acknowledgements}

The authors thank Aviral Kumar, Colin
Li, Katie Kang, Manan Tomar, Seohong Park, and the members of
RAIL for discussions and helpful feedback. The research was supported by the TPU Research Cloud, an NSF graduate fellowship, the Office of Naval Research (N00014-21-1-2838), ARO (W911NF-21-1-0097), AFOSR (FA9550-22-1-0273), and Intel.

\bibliography{example_paper}
\bibliographystyle{icml2022}

\newpage
\appendix
\onecolumn

\section{Didactic Example of ICVF}
\label{appendix:icvf_example}

Recall that the intention-conditioned value function (ICVF) is the function $V(s, s_+, z)$ with three inputs:
\begin{itemize}
    \item The current state of the agent $\mathbf{s}$
    \item A future outcome $\mathbf{s_+}$, which we would like to evaluate the likelihood of seeing in the future,
    \item An desired intention $\mathbf{z}$ (a task) that defines the policy being used to act.
\end{itemize}
Intuitively, we can read this expression as, 

\begin{center}
\textit{    ``If I start at state $\mathbf{s}$, what is the likelihood that I will see $\mathbf{s_+}$ in the process of acting to achieving some goal $\mathbf{z}$?''
}\end{center}

The formal definition of the ICVF is provided in Equation \ref{eq:icvf} as the expected discounted sum of indicator functions; this turns out to also have a nice probabilistic interpretation:

$$V(s, s_+, z) = \frac{1}{1-\gamma}P(s_T=s_+ | \pi_z, s_0=s), ~~~~~~~T \sim \text{Geom}(1-\gamma)$$

\begin{wrapfigure}{r}{0.3\linewidth}
  \begin{center}
    \vspace{-4em}
    \includegraphics[width=0.75\linewidth]{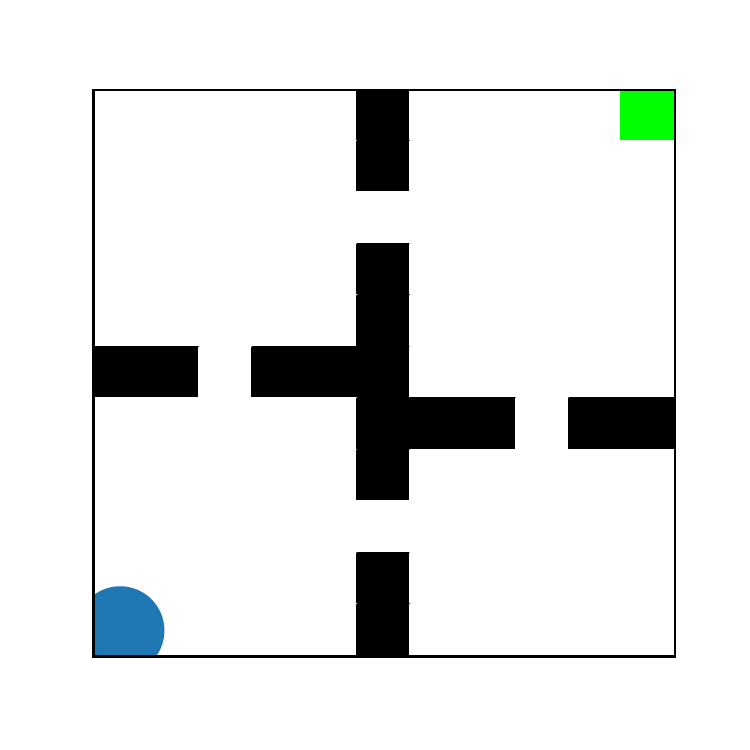}
  \end{center}
\end{wrapfigure}

\begin{figure}[H]
    \centering
    \includegraphics[width=\linewidth]{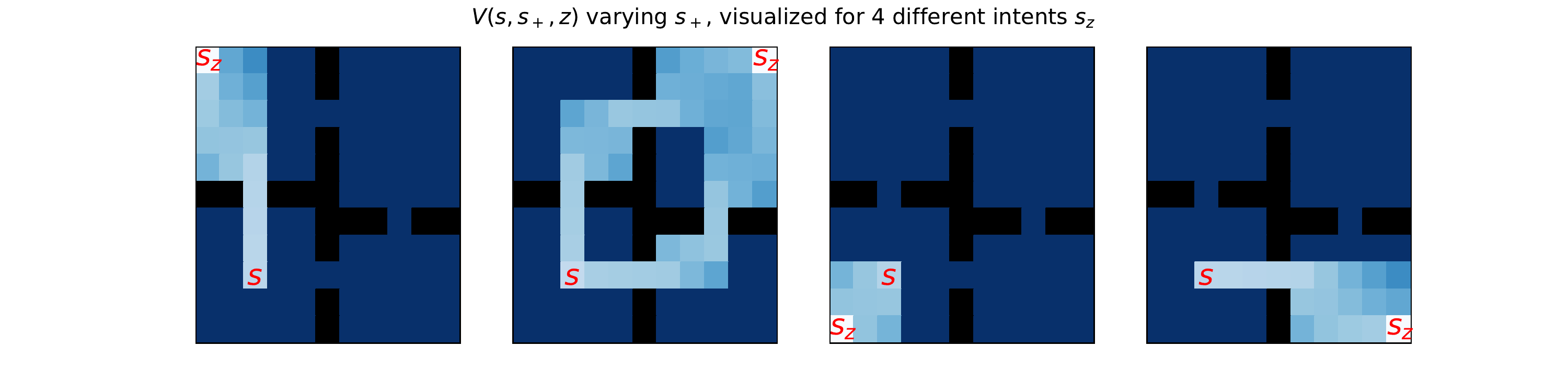}
    \caption{Using the ICVF to measure the state visitation distribution from a state $s$ when optimizing a task $z$: $V(s, s_+=\cdot, z)$. Visualized for $4$ different tasks.}
    \label{fig:example_icvf_probs}
\end{figure}

We visualize the ICVF in the Locked Doors domain, a discrete Gridworld environment (visualized on the right) where the intention-conditioned value function can be computed analytically. The first visualization, in Figure \ref{fig:example_icvf_probs}, plots the likelihood of seeing different outcome states, conditional on a specific state $s$ and a specific target $s_z$ marked in red. This means that the ICVF contains the information necessary to distinguish which states fall along the path from $s$ to $s_z$.

\begin{figure}[H]
    \centering
    \includegraphics[width=\linewidth]{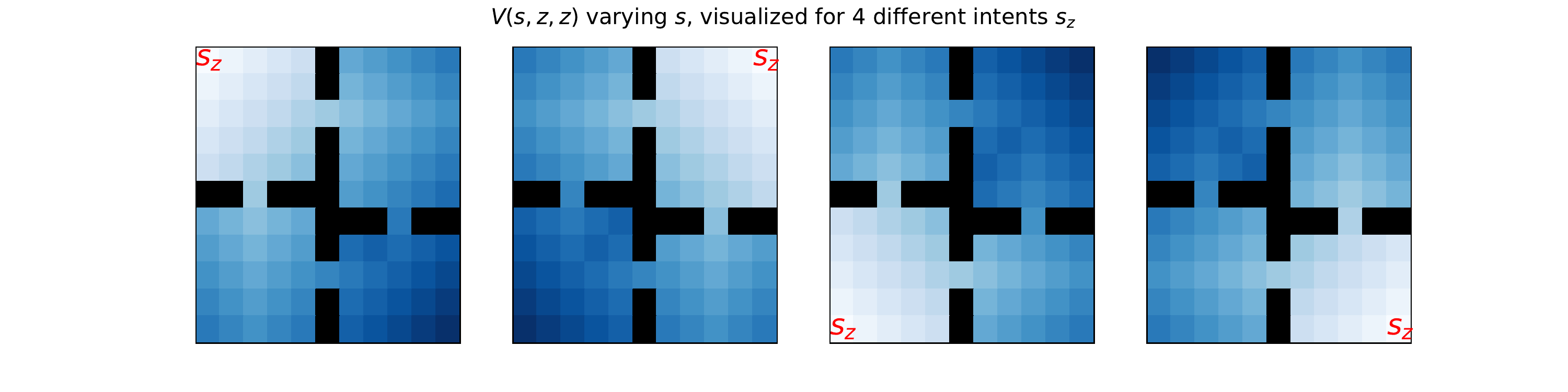}
    \caption{Using the ICVF to measure how likely an agent is to achieve the task $z$ when acting to achieve this task: $V(s=\cdot, z, z)$. Visualized for $4$ different tasks.}
    \label{fig:example_icvf_value}
\end{figure}

The ICVF, through Remark \ref{eq:icvf_representation}, is able to not only express the future state visitation probabilities, but also the value function for the tasks $z$ themselves. In Figure \ref{fig:example_icvf_value}, we plot the value functions for the intents derived from the ICVF, using the decomposition $V(s, z, z) = \sum_{s_+ \in \gS} r_z(s_+) V(s, s_+, z).$ This means that the ICVF can be used to also measure whether an agent is making progress towards some task / intention $z$.

\section{Proofs of Statements}

\begin{proposition}[Downstream value approximation]
    Suppose $\phi, \psi, T$ form an approximation to the true ICVF with error $\epsilon$, that is $\forall z \in \gZ$,
    $$\sum_{s, s_+ \in \gS}(V(s, s_+, z) - \phi(s)^\top T(z) \psi(s_+))^2 \leq \epsilon.$$
    For all rewards $r(s)$ and intentions $z \in \gZ$, $\exists \theta_r^z \in \R^d$ s.t.
    \begin{equation}\sum_{s \in \gS} (V_r^z(s) - \phi(s)^\top \theta_r^z)^2 \leq \epsilon \sum_{s_+ \in \gS} r(s_+)^2.
    \end{equation}
\end{proposition}

\begin{proof}
The proof follows immediately from the relationship between $V^*(s, s_+, z)$ and $V_r^z$. Let us write $V_r^z(s) = \sum_{s_+} r(s_+)V(s, s_+, z)$ and let $\theta = T(z) \sum_{s_+} r(s_+) \psi(s_+)$. 

\begin{align*}
    \sum_{s \in \gS} (V_r^z(s) - \phi(s)^\top \theta_r^z)^2  &= \sum_{s \in \gS} (\sum_{s_+} r(s_+)V(s, s_+, z) - \phi(s)^\top T(z) \sum_{s_+} r(s_+) \psi(s_+))^2 \\
     &= \sum_{s \in \gS} \left(\sum_{s_+} r(s_+) (V(s, s_+, z) - \phi(s)^\top T(z) \psi(s_+))\right)^2 \\
     \intertext{The result follows from an application of Cauchy-Schwarz}
     &\leq \sum_{s \in \gS} \big(\sum_{s_+} r(s_+)^2\big) \sum_{s_+}\big(V(s, s_+, z) - \phi(s)^\top T(z) \psi(s_+)\big)^2\\
     &= \big(\sum_{s_+} r(s_+)^2\big) \big(\sum_{s, s_+ \in \gS}  \big(V(s, s_+, z) - \phi(s)^\top T(z) \psi(s_+)\big)^2\big)\\
     &= \epsilon\left(\sum_{s_+} r(s_+)^2\right)
\end{align*}
\end{proof}

\section{Experimental Setup}
\label{appendix:exp_setup}

\subsection{General setup}

The general workflow for pre-training and finetuning is shared across all the domains and tasks. The representations are trained on passive data only for $N$ timesteps ($N$ varies by domain); after training, the learned parameters of the state representation are copied into the value network (if encoder parameters are shared between the value network and the actor or critic network, then the representation encoder parameters are also shared with this other network. 

For the finetuning stage, in our early experiments, we found that freezing the representation to the pre-trained representation often led to divergence in value estimation. To avoid this, in our finetuning setup, we allow for the representation to be finetuned, but add a constraint $\E_{\gD_{\text{downstream}}}[\gL_{\text{downstream}}] + \alpha \E_{\gD_{\text{pretrain}}}[\gL_{\text{pretrain}}]$ to ensure that the representation remains close to the original pre-trained representation. We did not experiment very closely with the choice of regularization, and it is feasible that simpler regularization penalties (e.g. $\ell_2$ penalties) may yield the same effect. We perform the coarse hyperparameter sweep $\alpha \in [1, 10, 100]$ for each representation and domain to choose $\alpha$, since the scales of the downstream losses can vary between domains, and the scales of the representation loss can vary between different methods. For our method, we set the expectile parameter for our method to $\alpha = 0.9$ for all the tasks, and train using temporal difference learning with a target network lagging via polyak averaging with rate $\lambda = 0.005$.

\begin{table}[]
\centering
\begin{tabular}{l|llll}
     & Antmaze & XMagical & Atari &  \\
     \hline
ICVF &   100      &    10      &  10     &  \\
VIP  &    10     &     10     &   10    &  \\
CPC  &   1      &      1    &   10    &  \\
CURL &   1      &       1   &   10    &  \\
APV  &   1000      &     1     &   10    & 
\end{tabular}
\caption{Choices for regularization constant $\alpha$ for each representation and domain}
\end{table}

\subsection{D4RL Antmaze Benchmark}

The D4RL Antmaze domain is an offline RL task that requires an ant quadruped agent to locomote in a maze environment to a target destination in the top right, by learning from an offline dataset of 1M frames of varying diversity. The state dimension is 29 and the action dimension is 8. Since we are interested in learning from passive data, we create a synthetic passive dataset by removing action and reward labels from the entire dataset, resulting in a dataset of observation sequences with 1 million transitions. For the downstream finetuning task, we create a smaller offline RL dataset by subsampling trajectories from the released offline dataset summing to $2.5 \times 10^5$ transitions (roughly 250 trajectories). No data is shared between different D4RL datasets.

Since the environment is state-based, we trained fully separate networks for each representation (e.g. no sharing between $\phi$ and $\psi$, modelling each as a 2 layer MLP with 256 hidden units each. Pre-training proceeds for 250k timesteps for all methods, which we found sufficient to lead to convergence for all the representation learning algorithms. We perform the downstream offline RL component using IQL \citep{kostrikov2021offline}, since the vanilla algorithm performs highly on the standard Antmaze tasks. During finetuning, we copy the pre-trained representation as the first $2$ layers of the value MLP $V(s)$ for IQL, leaving the network for the critic $Q(s, a)$ and for the policy un-altered. We use the hyperparameters for IQL from \citep{kostrikov2021offline}, running for 1M steps, and measuring final performance.

\subsection{XMagical Benchmark}

The \textbf{x-magical} suite is a benchmark extension of the MAGICAL Suite from \citet{toyer2020magical} testing cross-embodiment learning. This suite consists of one task (sweeping all three debris to the goal zone) and four different embodiments (Gripper, Shortstick, MediumStick, Longstick). The environment is from images, natively rendered at $384 \times 384$, although we subsample to $64 \times 64$. We build on the dataset of video released by \citet{zakka2021xirl}, which contains approximately 50000 transitions of each agent acting optimally. To increase the size of the video dataset, we additionally trained a BC policy on the released data (which is \textbf{not} optimal, unlike the original video data) and collect 100000 more transitions per agent. The finetuning task we set for the agent is to solve the task with the Gripper embodiment given access to 10000 actionful transitions sampled uniformly from these 150000 frames. We consider two passive data settings, first the \textbf{same-embodiment setting}, where the passive data consists of the 150000 frames from the Gripper embodiment, and the \textbf{cross-embodiment setting}, where the passive data joins together the frames from the ShortStick, Mediumstick, and Longstick embodiments. We use CQL \citep{kumar2020conservative} as the downstream offline RL algorithm, and use the Impala visual encoder \citep{Espeholt2018IMPALASD}. The encoder features are shared across the value, critic, and actor networks, but as in \citep{Kostrikov2020ImageAI}, do not allow the actor loss to modify the encoder weights.

\subsection{Atari 2600 with Youtube Videos}
\label{appendix:atari_details}

We use the same Atari 2600 setup as \citet{Agarwal2020AnOP}, focusing on the games Pong, Breakout, Seaquest, and QBert, -- please see there for full details about the experimental setup. The video dataset we pretrain on is a manually scraped collection of Youtube videos of these games; the protocol for scraping and processing these videos is described in the next subsection. For the finetuning phase, we use a uniform subsample of 100000 transitions from the replay buffer of a DQN agent released by \citet{Agarwal2020AnOP}. We take the same downstream offline RL algorithm as in \citet{kumar2020conservative}, a QRDQN agent with the CQL regularizer.

\subsubsection{Youtube dataset curation}
In this section, we discuss how we collect and process Youtube videos for our Atari experiments.

We selected videos of the Atari 2600 versions of our games, avoiding other versions such as Atari 7800. Included videos feature slight differences in color, angle, speed, and game play but we excluded videos with significant modifications (such as ball steering, breaking through multiple bricks, or invisible bricks in Breakout) or glitches (e.g. collision glitches in Seaquest). Videos were downloaded into webm, mkv, and mp4 formats.

When cropping, we attempted to include the same information and borders as the window in Atari gym environments and cut to include only game play. In our dataset, we inserted terminal flags at the end of each video but did not split up videos into episodes by inserting terminals at every restart.

In preprocessing frames, we imitated the strategy of the AtariPreprocessing gym wrapper with frame skipping, max pooling, gray scaling, and resizing to $84 \times 84$. For each video, we chose a frame skip of $1$, $2$, or $4$ frames to roughly match the frame rate of the Atari gym environment. As in Atari environments, observations stack 4 consecutive images at a time.

\begin{figure}
    \centering
    \includegraphics[width=0.6\linewidth]{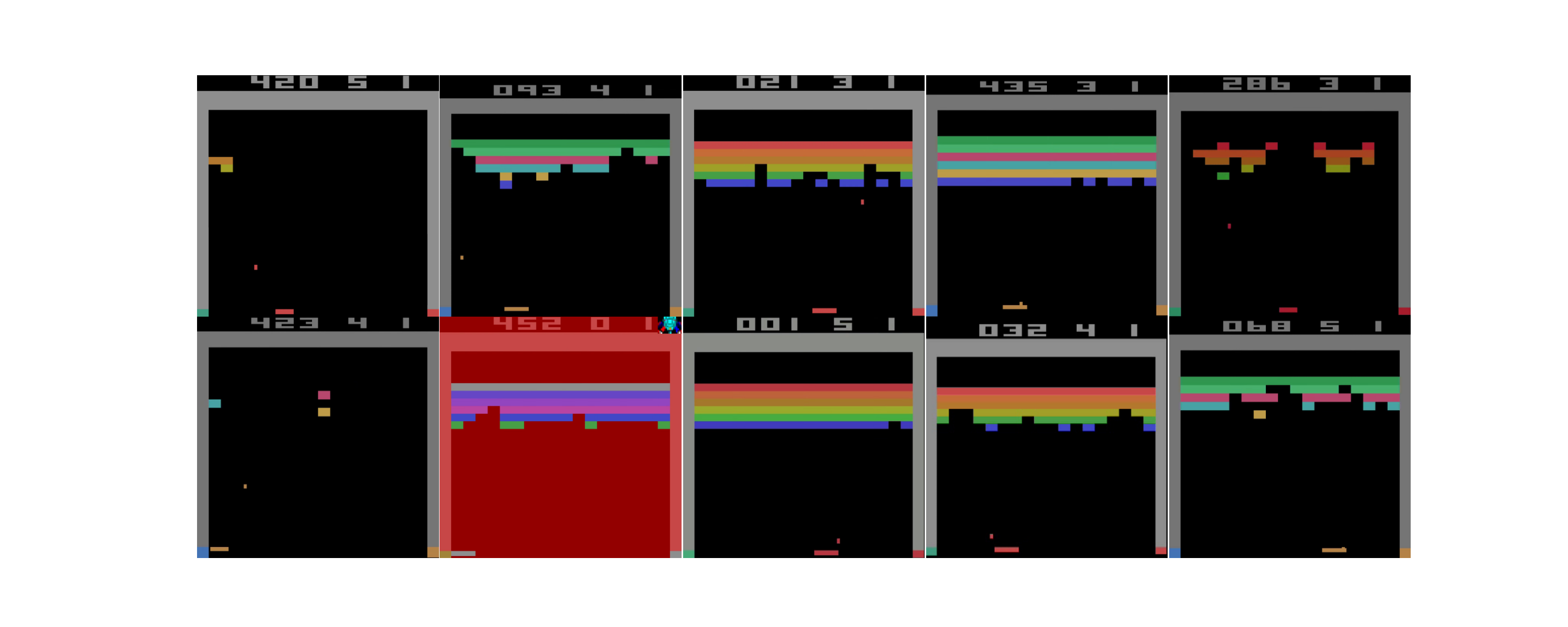}
    \includegraphics[width=0.3\linewidth]{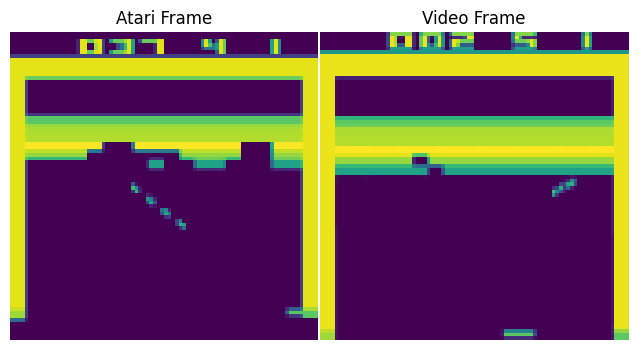}
    \caption{Breakout sample video frames shown above include color and speed differences, frame shifts, and even YouTube channel logos.}
    \label{fig:breakout_frames}
\end{figure}

\begin{figure}
    \centering
    \includegraphics[width=0.6\linewidth]{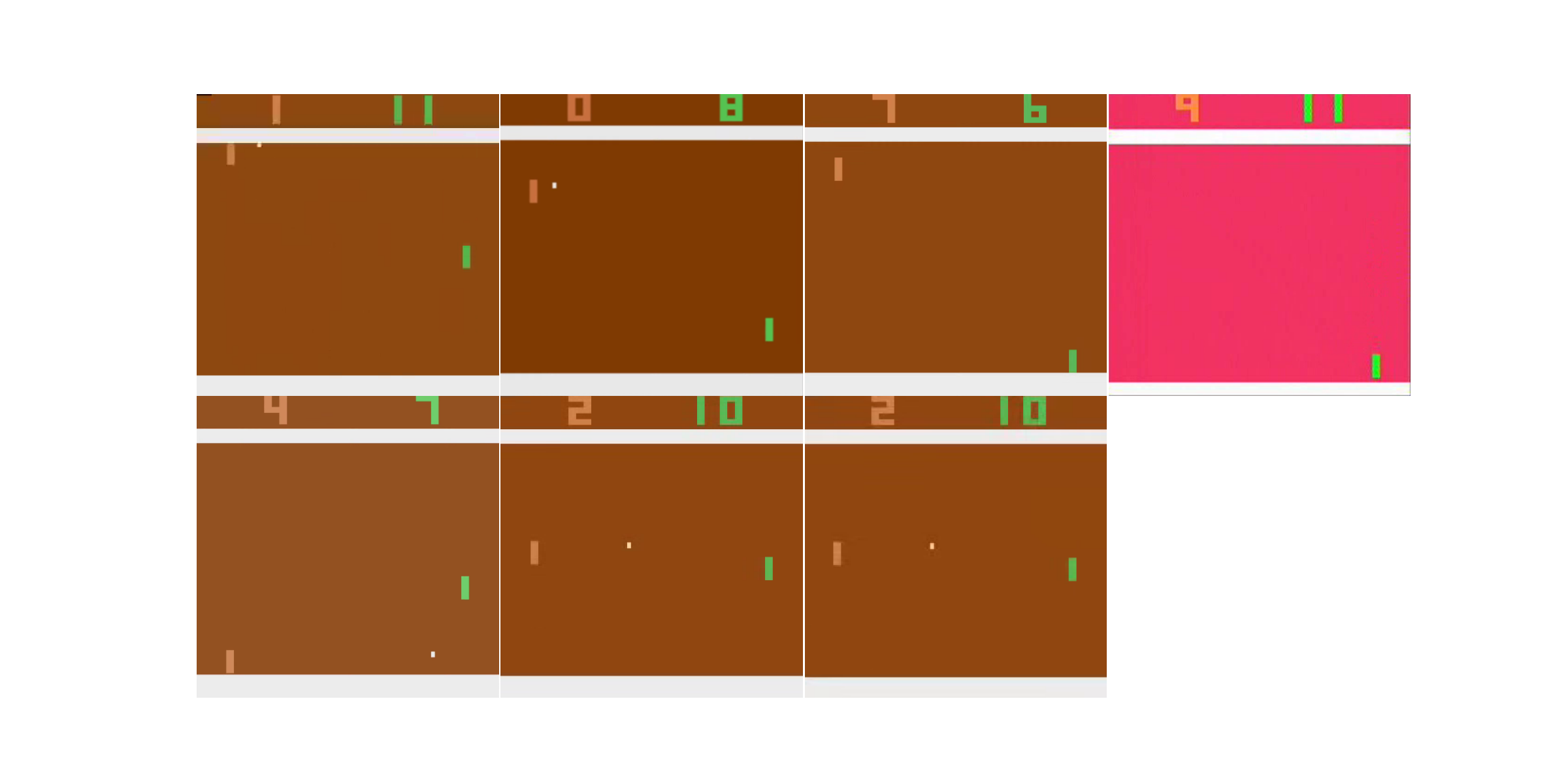}
    \includegraphics[width=0.3\linewidth]{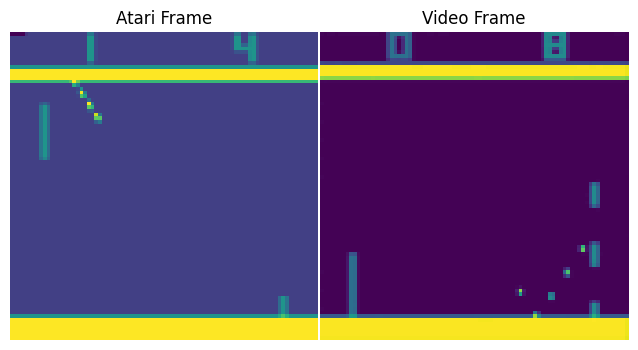}
    \caption{Pong sample video frames shown above feature color differences and speed differences.}
    \label{fig:pong_frames}
\end{figure}

\begin{figure}
    \centering
    \includegraphics[width=0.9\linewidth]{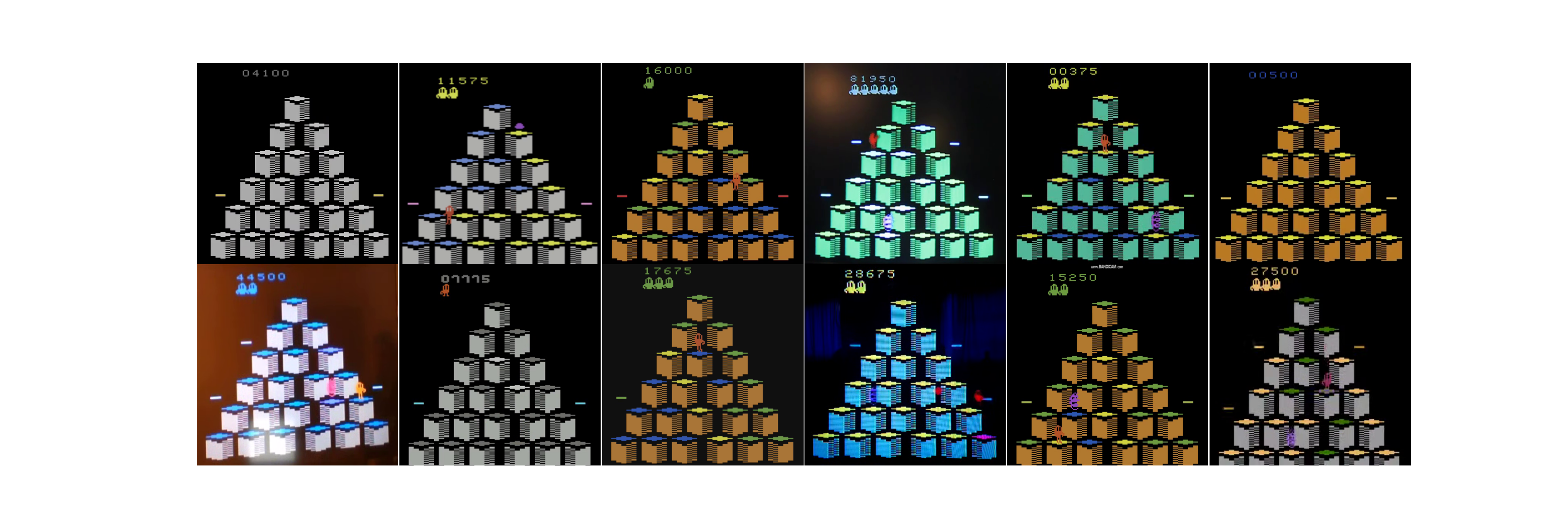}
    \caption{Qbert sample video frames shown above include angle and frame shifts, blur, font differences, and even reflections such as phone screens.}
    \label{fig:qbert_frames}
\end{figure}

\begin{figure}
    \centering
    \includegraphics[width=0.9\linewidth]{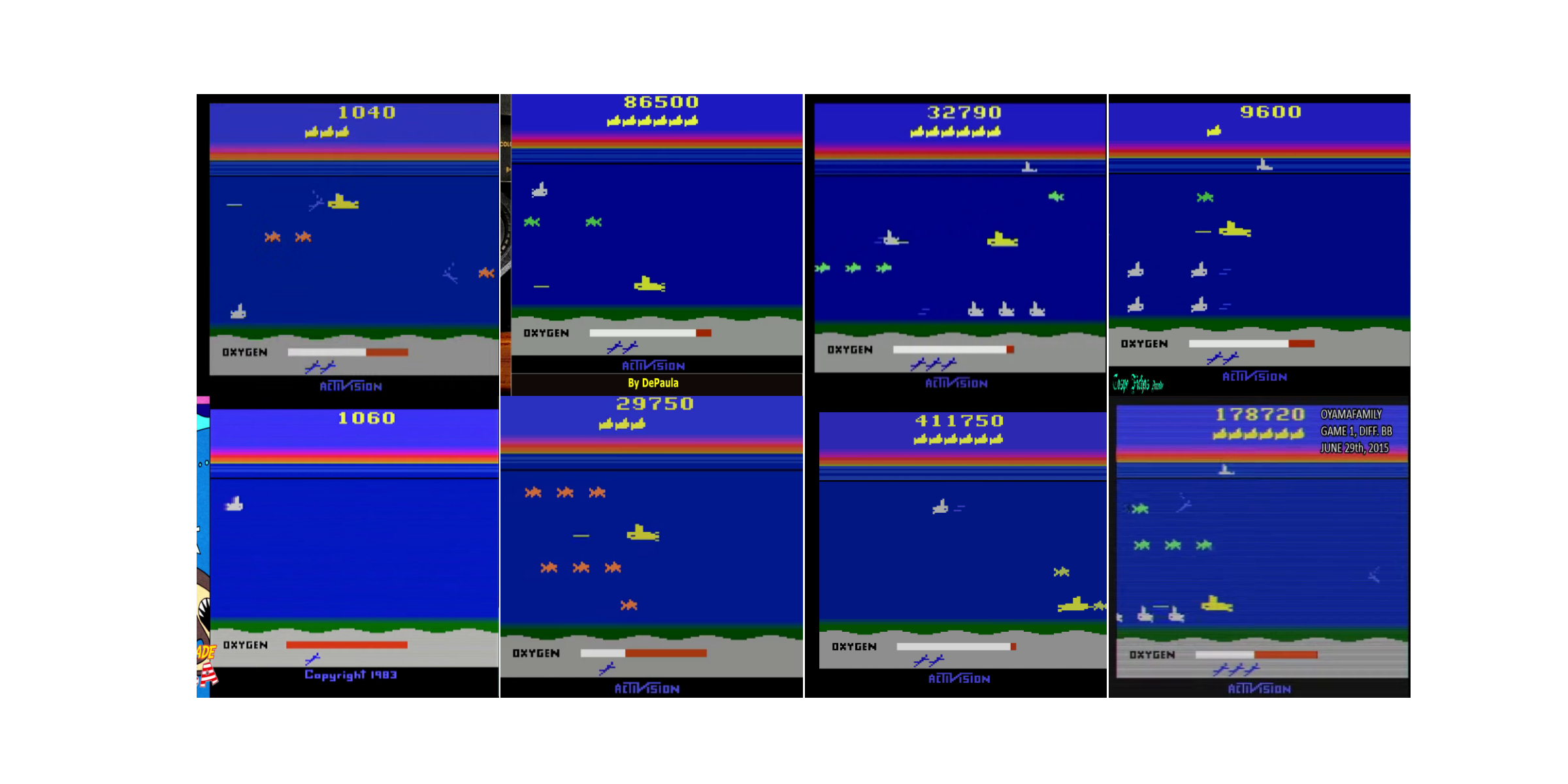}
    \caption{Seaquest sample video frames shown above include text overlays, logos, and frame shifts.}
    \label{fig:seaquest_frames}
\end{figure}

We show example frames in \ref{fig:breakout_frames}, \ref{fig:pong_frames}, \ref{fig:qbert_frames}, and \ref{fig:seaquest_frames}

\begin{center}
\begin{tabular}{l|c|c|r|r|r}
Game & Video Link & Clip Window & Start Frame & End Frame & Frame Skip\\
 & \texttt{https://youtu.be/...}& ($x_1,y_1,x_2,y_2$) & & \\
\hline
\hline
\footnotesize\textsc{Breakout}
& \href{https://youtu.be/Cr6z3AyhRr8}{\texttt{Cr6z3AyhRr8}} & 6,16,3077,2149 & 100 & 25650 & 2 \\
& \href{https://youtu.be/mBCr3zLn-p0}{\texttt{mBCr3zLn-p0}} & 2,27,1277,703 & 100 & 7250 & 2 \\
& \href{https://youtu.be/QcjDqSQrflU}{\texttt{QcjDqSQrflU}} & 6,39,1279,674 & 100 & 2500 & 2 \\
& \href{https://youtu.be/4CBjkD-Kbw8}{\texttt{4CBjkD-Kbw8}} & 4,38,1275,708 & 0 & 12950 & 2 \\
& \href{https://youtu.be/zluH_4eSmPs}{\texttt{zluH\_4eSmPs}} & 242,32,1680,996 & 100 & 9700 & 2 \\
& \href{https://youtu.be/Be6yDh_xKok}{\texttt{Be6yDh\_xKok}} & 2,52,1277,706 & 100 & 13441 & 2 \\
& \href{https://youtu.be/tT70Tv6D41o}{\texttt{tT70Tv6D41o}} & 5,5,2394,1428 & 3100 & 39800 & 2 \\
& \href{https://youtu.be/qx5dGzNVpyo}{\texttt{qx5dGzNVpyo}} & 96,8,1189,664 & 250 & 2550 & 1 \\
& \href{https://youtu.be/1oATuwG_dsA}{\texttt{1oATuwG\_dsA}} & 68,70,576,388 & 300 & 3100 & 2 \\
& \href{https://youtu.be/IJWrq01B3gM}{\texttt{IJWrq01B3gM}} & 4,43,1273,703 & 50 & 5250 & 2 \\
\hline
\footnotesize\textsc{Pong}
& \href{https://youtu.be/p88R2_3yWPA}{\texttt{p88R2\_3yWPA}} & 1,1,398,222 & 50 & 3985 & 1 \\
& \href{https://youtu.be/XmZxUl0k8ew}{\texttt{XmZxUl0k8ew}} & 215,12,1707,1079 & 150 & 4700 & 1 \\
& \href{https://youtu.be/YBkEnBqZBFU}{\texttt{YBkEnBqZBFU}} & 1,32,639,449 & 100 & 9600 & 1 \\
& \href{https://youtu.be/CxgaZenhngU}{\texttt{CxgaZenhngU}} & 34,77,686,414 & 50 & 7500 & 1 \\
& \href{https://youtu.be/jiFCgAcM2tU}{\texttt{jiFCgAcM2tU}} & 259,7,1663,1079 & 200 & 11200 & 1 \\
& \href{https://youtu.be/moqeZusEMcA}{\texttt{moqeZusEMcA}} & 0,1,159,208 & 50 & 3447 & 1 \\
& \href{https://youtu.be/T9zJmcR047w}{\texttt{T9zJmcR047w}} & 1,1,159,207 & 50 & 3450 & 1 \\
\hline
\footnotesize\textsc{Qbert}
& \href{https://youtu.be/QVuAJFDydDU}{\texttt{QVuAJFDydDU}} & 141,5,1118,715 & 50 & 12600 & 4 \\
& \href{https://youtu.be/pAK9-l2mLAA}{\texttt{pAK9-l2mLAA}} & 4,4,1918,1076 & 500 & 2700 & 4 \\
& \href{https://youtu.be/PdnYB9o3IWU}{\texttt{PdnYB9o3IWU}} & 0,1,553,359 & 50 & 5200 & 4 \\
& \href{https://youtu.be/AnKkbQh6a0k}{\texttt{AnKkbQh6a0k}} & 69,4,594,415 & 0 & 56350 & 4 \\
& \href{https://youtu.be/W-gTlZYWkYM}{\texttt{W-gTlZYWkYM}} & 0,1,545,360 & 0 & 1852 & 4 \\
& \href{https://youtu.be/Jq0DGEoZEQY}{\texttt{Jq0DGEoZEQY}} & 452,27,1178,576 & 3100 & 12150 & 4 \\
& \href{https://youtu.be/zGt-i0cL1tQ}{\texttt{zGt-i0cL1tQ}} & 171,36,663,328 & 600 & 46748 & 4 \\
& \href{https://youtu.be/DhzgJQF_wuU}{\texttt{DhzgJQF\_wuU}} & 2,1,493,358 & 0 & 3625 & 4 \\
& \href{https://youtu.be/vkZhWsiHCqM}{\texttt{vkZhWsiHCqM}} & 144,7,1784,1076 & 150 & 8250 & 4 \\
& \href{https://youtu.be/CeTClJT5BvU}{\texttt{CeTClJT5BvU}} & 76,41,548,413 & 50 & 12037 & 4 \\
& \href{https://youtu.be/opQ6qdAee0U}{\texttt{opQ6qdAee0U}} & 4,5,1279,701 & 150 & 5650 & 4 \\
& \href{https://youtu.be/_tjsVgdAxFY}{\texttt{\_tjsVgdAxFY}} & 21,51,462,359 & 0 & 18897 & 4 \\
\hline
\footnotesize\textsc{Seaquest}
& \href{https://youtu.be/A9GNDwad27E}{\texttt{A9GNDwad27E}} & 68,72,574,392 & 150 & 3450 & 2 \\
& \href{https://youtu.be/Ir1t57ItCpw}{\texttt{Ir1t57ItCpw}} & 83,3,1160,719 & 350 & 44800 & 2 \\
& \href{https://youtu.be/V2ShNbzhN_E}{\texttt{V2ShNbzhN\_E}} & 176,45,1122,655 & 650 & 6550 & 2 \\
& \href{https://youtu.be/j3d1eZGqDoU}{\texttt{j3d1eZGqDoU}} & 2,1,1279,717 & 50 & 10350 & 2 \\
& \href{https://youtu.be/QK9wa-6hd6g}{\texttt{QK9wa-6hd6g}} & 155,7,1688,1079 & 8600 & 16800 & 2 \\
& \href{https://youtu.be/ZeRGJk7HQGc}{\texttt{ZeRGJk7HQGc}} & 1,2,483,343 & 300 & 13150 & 2 \\
& \href{https://youtu.be/PBsLzxvYUYM}{\texttt{PBsLzxvYUYM}} & 144,15,2877,2145 & 150 & 61500 & 2 \\
& \href{https://youtu.be/Y1yASa4j2wk}{\texttt{Y1yASa4j2wk}} & 21,20,639,460 & 250 & 60800 & 2 \\
\end{tabular}
\end{center}

\subsection{Video Augmentation}
\label{appendix:video_augmentation}
To increase the quantity of video data and bridge the gap to Atari environment frames, we randomly augment our video frames via random crops, contrast jittering, and brightness jittering. After padding frames by a pad width $p$, we crop $U(0, 2p)$ pixels off the width and height dimensions. Then, we multiply demeaned images by uniform random contrast jitter, add uniform random brightness jitter, and add the resulting noise to the original image.

\clearpage
\section{Additional Experiments}

\subsection{D4RL Antmaze}
\label{appendix:d4rl_extra}
\begin{figure}
    \centering
    \includegraphics[width=\linewidth]{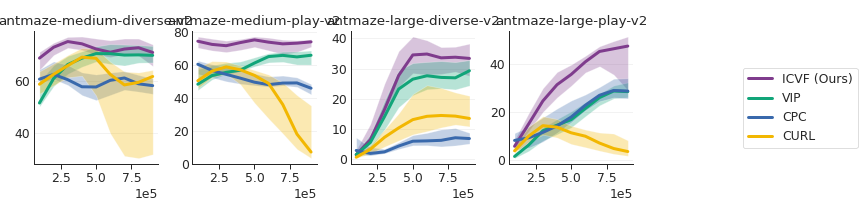}
    \caption{Performance curves through downstream finetuning on all Antmaze environments.}
    \label{fig:antmaze_graph}
\end{figure}

\textbf{Analysis:}
\begin{wrapfigure}{r}{0.5\linewidth}
  \begin{center}
    \includegraphics[width=0.75\linewidth]{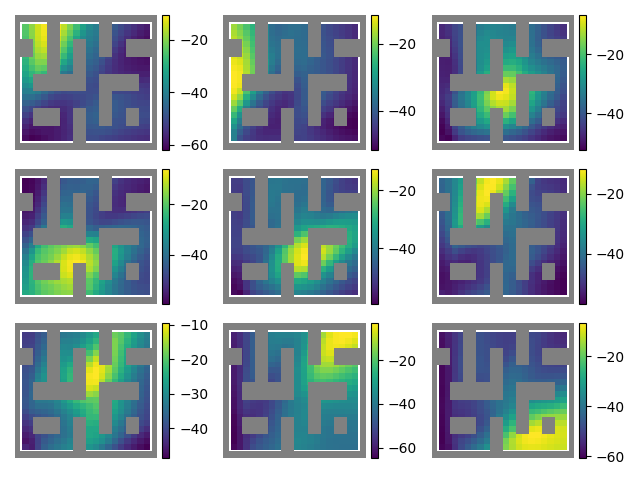}
  \end{center}
    \caption{Visualization of $\hat{V}(s, z, z)$ for different sampled intentions $z$ in the Antmaze.}
    \label{fig:intent_viz}
\end{wrapfigure}
We performed a simple analysis of the ICVF representation in the Antmaze domain, examining three components: 1) how well it can represent different downstream value functions, 2) what aspects of the state it attends to, and 3) what ICVF value functions look like for different sampled latent intentions $z$. In the table below, we measure how well the ICVF representation can represent the optimal value functions for various goal-reaching rewards. We learn value functions by performing SARSA with a linear head on top of a frozen representation, and measure absolute value error ($|V^* - \hat{V}|$) and gradient alignment ($(\nabla_s V^*)^\top (\nabla_s \hat{V}) > 0$). To examine which components of state the ICVF focuses on, we train a decoder to reconstruct the original state from the representation $\phi(s)$. The ICVF representation primarily captures the XY coordinates of the Ant (2\% relative error), with more lossy representations of joint angles (15\% error) or joint velocities (40\% error). Finally, in Figure \ref{fig:intent_viz}, we visualize the ICVF value functions for sampled intentions during the training process.

\begin{table}[!ht]
    \centering
    \begin{tabular}{|l|l|l|l|}
    \hline
        ~ & $|V^*-V$| & Value Gradient Alignment & Value Gradient Alignment near Goal \\ \hline
        ICVF & 11.9 $\pm$ 0.9 & 74\% $\pm$ 2\% & 81\% $\pm$ 2\% \\ \hline
        CURL & 12.9 $\pm$ 1.0 & 64\% $\pm$ 1\% & 73\% $\pm$ 2\% \\ \hline
        VIP & 13.8 $\pm$ 1.1 & 70\% $\pm$ 2\% & 78\% $\pm$ 2\% \\ \hline
        CPC & 12.9  $\pm$ 0.9 & 63\% $\pm$ 1\% & 71\% $\pm$ 1\% \\ \hline
    \end{tabular}
    \caption{Accuracy of value functions finetuned on each representation, averaged across 10 randomly sampled goal-reaching tasks.}
\end{table}

\begin{table}[!ht]
    \centering
    \begin{tabular}{|l|l|l|l|l|}
    \hline
        ~ & $d=4$ & $d=32$ & $d=256$ & $d=2048$ \\ \hline
        \texttt{antmaze-medium-diverse-v2} & 0.48 $\pm$ 0.12 & 0.64 $\pm$ 0.08 & 0.66 $\pm$ 0.06 & 0.68 $\pm$ 0.08 \\ \hline
        \texttt{antmaze-large-diverse-v2} & 0.02 $\pm$ 0.02 & 0.20 $\pm$ 0.10 & 0.32 $\pm$ 0.11 & 0.3 $\pm$ 0.10 \\ \hline
    \end{tabular}
    \caption{Ablating dimensionality of the representation in the multilinear ICVF}
\end{table}

\textbf{}

\clearpage
\subsection{Example XMagical ICVF}

\begin{figure}[H]
    \centering
    \includegraphics[width=\linewidth]{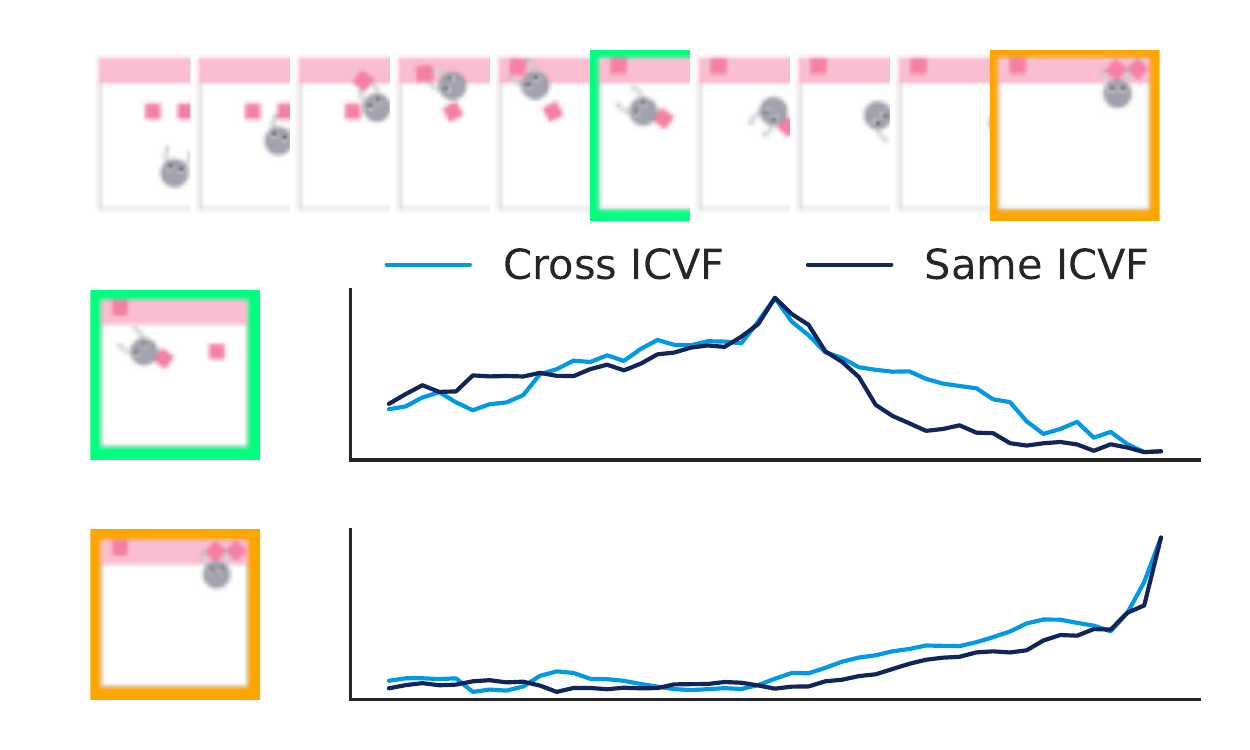}
    \caption{Comparison of learned ICVFs on XMagical data in the same embodiment versus the one trained on a different embodiment. Notice that despite being trained on different embodiments, the cross-embodiment ICVF nonetheless learns a good value function.}
    \label{fig:example_icvf2}
\end{figure}

\subsection{Example Atari ICVF}

\label{appendix:atari_extra}

\begin{figure}[H]
    \centering
    \includegraphics[width=\linewidth]{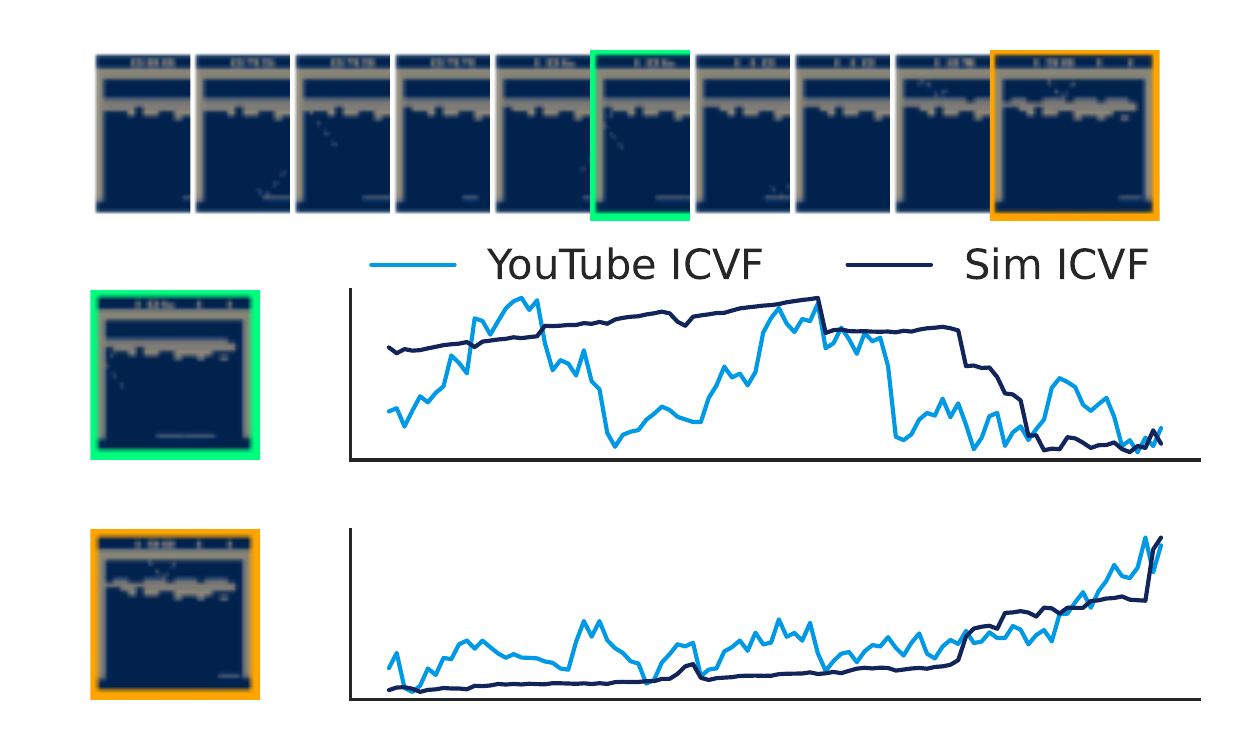}
    \caption{Comparison of learned ICVFs on Atari YouTube data vs. on Atari environment data. We train two ICVFs, one on simulated data and the other on Youtube video, and evaluate $V(s, z, z)$ for the goal-reaching tasks to the highlighted frames and show the results in \ref{fig:example_icvf}. The ICVF trained with Youtube data is not locally smooth, but has the correct global structure, indicating that the representation is able to correctly model long-horizon components.}
    \label{fig:example_icvf}
\end{figure}

\end{document}